\def\year{2021}\relax
\documentclass[letterpaper]{article} 

\usepackage{pdfpages}

\usepackage{aaai21_arxiv}  
\usepackage{times}  
\usepackage{helvet} 
\usepackage{courier}  
\usepackage[hyphens]{url}  
\usepackage{graphicx} 
\usepackage{natbib}  
\usepackage{caption} 
\usepackage[utf8]{inputenc} 

\usepackage{physics}
\usepackage{mathtools}
\usepackage{verbatim}
\usepackage{xcolor}
\usepackage{xr}

\usepackage{algorithm}
\usepackage{algorithmic}

\newcommand{\RR}{I\!\!R} 

\usepackage{placeins}
\usepackage{booktabs}
\usepackage{multirow}
\usepackage{bbold}

\usepackage{subcaption} 

\title{Online Spatio-Temporal Learning in Deep Neural Networks}

\author{
    Thomas Bohnstingl\textsuperscript{\rm 1, \rm 2}\thanks{Correspondance to boh@zurich.ibm.com},
    Stanisław Woźniak\textsuperscript{\rm 1}, 
    Wolfgang Maass\textsuperscript{\rm 2}, \\
    Angeliki Pantazi\textsuperscript{\rm 1},
    Evangelos Eleftheriou\textsuperscript{\rm 1} \\
}
\affiliations{
    \textsuperscript{\rm 1}IBM Research – Zurich\\
    8803 Rüschlikon\\
    \textsuperscript{\rm 2}Institute of Theoretical Computer Science\\
    Graz University of Technology\\
}

\begin{document}
\maketitle



\begin{abstract}
Biological neural networks are equipped with an inherent capability to continuously adapt through online learning. This aspect remains in stark contrast to learning with error backpropagation through time (BPTT) applied to recurrent neural networks (RNNs), or recently to biologically-inspired spiking neural networks (SNNs). BPTT involves offline computation of the gradients due to the requirement to unroll the network through time. Online learning has recently regained the attention of the research community, focusing either on approaches that approximate BPTT or on biologically-plausible schemes applied to SNNs. Here we present an alternative perspective that is based on a clear separation of spatial and temporal gradient components. Combined with insights from biology, we derive from first principles a novel online learning algorithm for deep SNNs, called online spatio-temporal learning (OSTL). For shallow networks, OSTL is gradient-equivalent to BPTT enabling for the first time online training of SNNs with BPTT-equivalent gradients. In addition, the proposed formulation unveils a class of SNN architectures trainable online at low time complexity. Moreover, we extend OSTL to a generic form, applicable to a wide range of network architectures, including networks comprising long short-term memory (LSTM) and gated recurrent units (GRU). We demonstrate the operation of our algorithm on various tasks from language modelling to speech recognition and obtain results on par with the BPTT baselines. The proposed algorithm provides a framework for developing succinct and efficient online training approaches for SNNs and in general deep RNNs.
\end{abstract}

\newcommand{\figOne}[1]{
\begin{figure*}[#1]
\centering
\includegraphics{./BPTT_OSTL_boh.png}
\caption{\textbf{Gradient flow for a single layer RNN unfolded for three time steps}. The parameter update at time step $t$ requires all previous activities of all layers for $1 \leq t' \leq t$ and the gradients from the loss function need to propagate back in time until the beginning of the input sequence.}
\label{fig:MultiLayerRNNBPTT}
\end{figure*}
}

\newcommand{\figTwo}[1]{
\begin{figure}[#1]
\centering
\includegraphics[width=0.85\columnwidth]{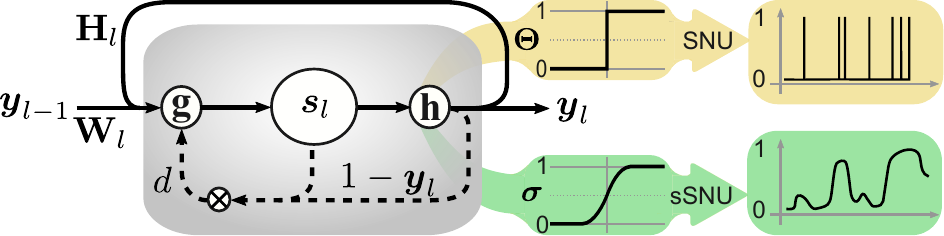}
\caption{\textbf{SNU} incorporates LIF dynamics in an RNN unit.}
\label{fig:SNU}
\end{figure}
}

\newcommand{\figThree}[1]{
\begin{figure*}[#1]
\centering
\includegraphics[width=1.87\columnwidth]{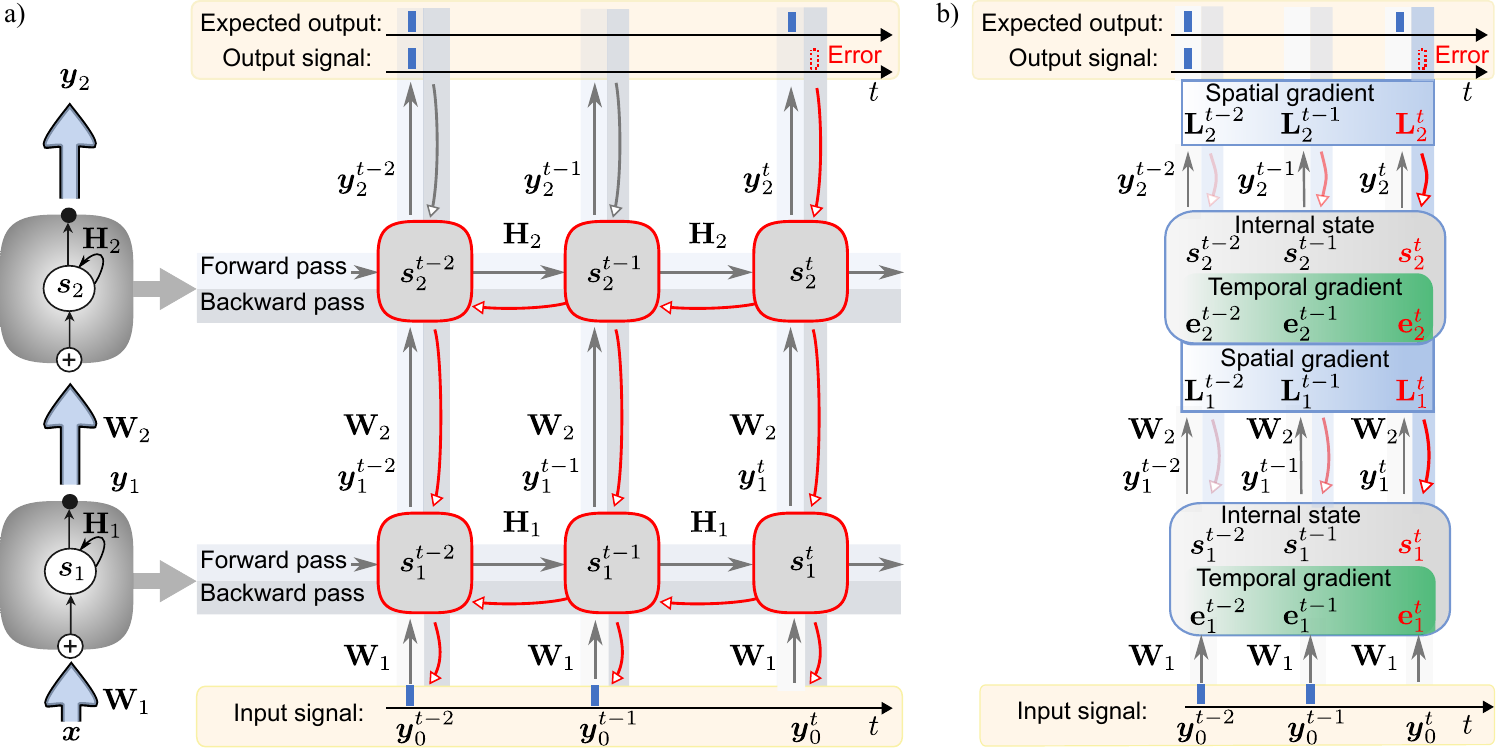}
\caption{\textbf{Conceptual illustration of the gradient flow of a two-layered generic recurrent neural network}. a) In BPTT, the unfolded network components involved in the gradient computation at time step $t$ are marked in red. b)~In OSTL, the spatial and temporal components are clearly separated. Each layer computes eligibility traces which account for the temporal gradients. Independently, an individual learning signal per layer passes from the output layer through the network and accounts for the spatial components. The components involved in the gradient computation at time $t$ are marked in red.}
\label{fig:MultiLayerGradientFlowRNN}
\end{figure*}
}

\newcommand{\figFour}[1]{
\begin{figure*}[#1]
\centering
\includegraphics[width=0.45\columnwidth]{./Results_JSB.png}
\caption{\textbf{Test performance on JSB dataset} for sSNU and SNU trained with BPTT (blue bar) and OSTL (orange bar). The negative log-likelihood loss was used -- lower value is better.}
\label{fig:Results_JSB}
\end{figure*}
}

\newcommand{\figFive}[1]{
\begin{figure*}[#1]
\centering
\includegraphics[width=0.6\columnwidth]{./Results_RC_MNIST_SNU.png}
\caption{\textbf{Comparison for the rate-coded MNIST dataset}. Test performance for different SNUs trained with BPTT (blue line) and OSTL (orange line) for the rate-coded MNIST dataset. The classification accuracy was used -- higher value is better.}
\label{fig:Results_RCM_MNIST}
\end{figure*}
}

\newcommand{\tabOne}[1]{
\begin{table}[#1] 
\caption{Performance comparison of OSTL and BPTT.\\
\,*State-of-the-art results from literature~\cite{Wozniak2020Jun}}
\label{performanceComparisonTable}
\vskip 0.15in
\begin{center}
\begin{small}
\begin{sc}
\begin{tabular}{llr@{}lr@{}l}
Task & RNN unit & \multicolumn{2}{c}{OSTL} & \multicolumn{2}{c}{BPTT} \\
\midrule
JSB & sSNU & \textbf{8.40\,} &\textbf{$\pm$\, 0.02} & *8.39\, &$\pm$\, 0.01 \\
JSB & SNU & \textbf{8.72\,} &\textbf{$\pm$\, 0.02} & *8.72\, &$\pm$\, 0.04 \\
RC-MNIST & sSNU & \textbf{96.33\,} &\textbf{$\pm$\, 0.10} & 98.22\, &$\pm$\, 0.05 \\
RC-MNIST & SNU & \textbf{95.34\,} &\textbf{$\pm$\, 0.24} & 97.79\, &$\pm$\, 0.09 \\
PTB & sSNU & \textbf{138.3ppl\,} &\textbf{$\pm$\, 0.5} & *137.7ppl\, &$\pm$\, 1.6\\
\bottomrule
\end{tabular}
\end{sc}
\end{small}
\end{center}
\end{table}
}

\newcommand{\tabTwo}[1]{
\begin{table*}[#1] 
\caption{Comparison of OSTL and algorithms from literature. The complexities correspond to a single parameter update in single-layer neural networks unless otherwise indicated.}

\label{complexityComparisonTable}
\begin{center}
\begin{small}
\begin{sc}
\begin{tabular}{lcccc}
\toprule
Algorithm & 
\shortstack{Memory \\complexity} & 
\shortstack{Time \\complexity} & 
\shortstack{Exact gradients\\(vs. BPTT)} &
Derived for \\
\midrule
BPTT (unrolled for $T$ time steps) & $T n$ & $T n^2$ & $\surd$ & RNNs\\
RTRL with deferred updates & $n^3$ & $n^4$ & $\surd$ & RNNs\\
RTRL & $n^3$ & $n^4$ & $\times$ & RNNs\\
\hline\\[-2.1ex]
UORO & $n^2$ & $n^2$ & $\times$ & RNNs\\
KF-RTRL & $n^2$ & $n^3$ & $\times$ & RNNs\\
OK-RTRL (for $r$ summation terms) & $r n^2$ & $r n^3$ & $\times$ & RNNs\\
RFLO & $n^2$ & $n^2$ & $\times$ & SNNs \& RNNs\\
SuperSpike (for integration period $T$) & $n^2$ & $T n^2$ & $\times$ & SNNs\\
E-prop & $n^2$ & $n^2$ & $\times$ & SNNs \& RNNs\\
\hline\\[-2.1ex]
\textbf{OSTL: feed-forward SNNs ($k$ layers)} & ${kn^2}$ & ${kn^2}$ & $\times$ & SNNs\\

\textbf{OSTL: recurrent SNNs ($k$ layers, w/o $\mathbf{H}$)} & ${kn^2}$ & ${kn^2}$ & $\times$ & SNNs\\
\textbf{OSTL: feed-forward SNNs} & ${n^2}$ & ${n^2}$ & $\surd$ & SNNs\\
\textbf{OSTL: recurrent SNNs (w/o $\mathbf{H}$)} & ${n^2}$ & ${n^2}$ & $\times$ & SNNs\\
\textbf{OSTL: generic RNNs ($k$ layers)} & ${k n^3}$ & ${k n^4}$ & $\times$ & SNNs \& RNNs\\
\textbf{OSTL: generic RNNs} & ${n^3}$ & ${n^4}$ & $\surd$ &  SNNs \& RNNs
\end{tabular}
\end{sc}
\end{small}
\end{center}
\end{table*}
}

\newcommand{\tabThree}[1]{
\begin{table}[#1] 
\caption{Performance comparison for the TIMIT dataset. $T$ -- truncation horizon of TBPTT.}
\label{performanceComparisonTIMIT}
\vskip 0.15in
\begin{center}
\begin{small}
\begin{sc}
\begin{tabular}{lllr@{}l}
\toprule
RNN unit & Algorithm & \multicolumn{2}{c}{Error rate}\\
\midrule
LSTM & BPTT & 19.63\%\, & $\pm$\, 0.25 \\
LSTM & TBPTT ($T=90$) & 19.78\%\, & $\pm$\, 0.11 \\
LSTM & TBPTT ($T=50$) & 20.56\%\, & $\pm$\, 0.11 \\
LSTM & TBPTT ($T=30$) & 21.34\%\, & $\pm$\, 0.09 \\
LSTM & TBPTT ($T=10$) & 28.61\%\, & $\pm$\, 0.68 \\
LSTM & OSTL & \textbf{20.35\%\,} &\textbf{$\pm$\, 0.11} \\
sSNU & BPTT & \textbf{20.90\%\,} &\textbf{$\pm$\, 0.10}\\
sSNU & OSTL & \textbf{21.60\%\,} &\textbf{$\pm$\, 0.10}\\
\bottomrule
\end{tabular}
\end{sc}
\end{small}
\end{center}
\vskip -0.1in
\end{table}
}

\newcommand{\figJoint}[1]{
\begin{figure*}[#1]
\centering
\includegraphics[width=2.0\columnwidth]{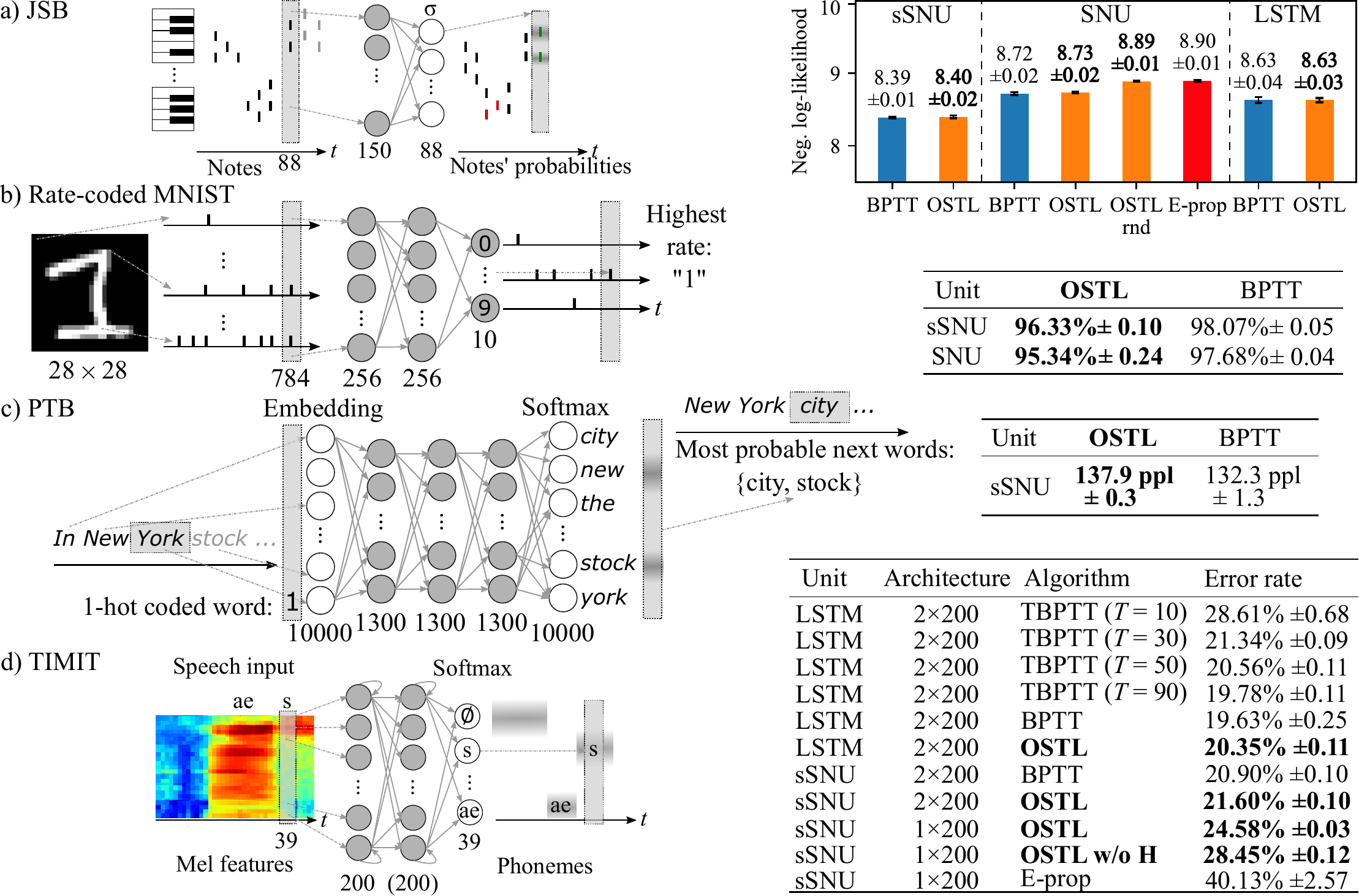}
\caption{\textbf{Tasks and test performance} For each task a schematic depiction of the input and the network architecture is presented on the left side and the results on the right side. RNN units are drawn in grey and stateless units in white. OSTL results in bold font. a) JSB dataset trained with BPTT (blue bars), with random E-prop (red bar) and with OSTL (orange bars). The negative log-likelihood loss was used  as a performance metric (lower value is better).
b)~Rate-coded MNIST dataset. We used the classification accuracy for assessment. c) PTB dataset. The perplexity was used as the performance metric (lower value is better). d)~TIMIT dataset. $T$ -- truncation horizon of TBPTT. For the assessment, the error rate was used (lower value is better).}
\label{fig:Results}
\end{figure*}
}

\newcommand{\algoOne}[1]{
\begin{algorithm}[#1]
\caption{OSTL for deep FF-SNNs with MSE loss}
\begin{algorithmic}
\label{alg:OSTL}
\STATE {\bfseries Input:} $(\boldsymbol{\theta}_l)_{l=1,\dots,k}$,  $\eta$, data ($\boldsymbol{y}_0^t,\hat{\boldsymbol{y}_k}^t)_{t=1,\dots,T}$, initialization: e.g. $(\boldsymbol{s}_l^0=\boldsymbol{y}_l^0=\boldsymbol{\epsilon}_l^{0,\boldsymbol{\theta}}=\Delta\boldsymbol{\theta}_l=0)_{l=1,\dots,k}$.
\STATE {\bfseries Output:} Parameter update $(\Delta \boldsymbol{\theta}_l)_{l=1,\dots,k}$
 \FOR{$1 \leq t \leq T$}
 \FOR{$1 \leq l \leq k$}
  \STATE Compute $\boldsymbol{s}_l^t$ using Eq.~\ref{eq:SNUState} with $\boldsymbol{y}_{l-1}^t,\boldsymbol{s}_l^{t-1},\boldsymbol{y}_l^{t-1},\boldsymbol{\theta}_l$
  \STATE Compute $\boldsymbol{y}_l^t$ using Eq.~\ref{eq:SNUOutput} with $\boldsymbol{s}_l^{t},\boldsymbol{\theta}_l$
  \STATE Compute $\boldsymbol{\epsilon}_l^{t,\boldsymbol{\theta}}$ using Eq.~\ref{eq:FFExplicitEpsW}/\ref{eq:FFExplicitEpsb} with $\boldsymbol{\epsilon}_l^{t-1,\boldsymbol{\theta}},\boldsymbol{s}_l^t,\boldsymbol{y}_l^t,\boldsymbol{\theta}_l$
  \STATE Compute $\mathbf{e}_l^{t,\boldsymbol{\theta}}$ using Eq.~\ref{eq:FFExplicitEW}/\ref{eq:FFExplicitEb} with $\boldsymbol{\epsilon}_l^{t,\boldsymbol{\theta}},\boldsymbol{s}_l^t,\boldsymbol{y}_l^t,\boldsymbol{\theta}_l$
  \ENDFOR
  \STATE Compute $E^t$ using Eq.~\ref{eq:MeanSquaredLossFunction} with $\boldsymbol{y}_k^t,\hat{\boldsymbol{y}_k}^t$
  \FOR{$1 \leq l \leq k$}
  \STATE Compute $\mathbf{L}_l^t$ using Eq.~\ref{eq:MeanSquaredLossFunction}
  \STATE Compute $\mathrm{d}E^t=-\eta \left(\mathbf{L}_l^t \mathbf{e}_l^{t,\boldsymbol{\theta}}\right)$\\
  \STATE Update $\boldsymbol{\theta}_l$ with 
  $\mathrm{d}E^t \left[\mathrm{Online\,updates}\right]$\\
  \STATE Accumulate $\Delta\boldsymbol{\theta}_l=\Delta\boldsymbol{\theta}_l 
  +\mathrm{d}E^t \left[\mathrm{Deferred\,updates}\right]$
\ENDFOR
\ENDFOR
\STATE Update $\boldsymbol{\theta}_l$ with accumulated $\Delta\boldsymbol{\theta}_l \left[\mathrm{Deferred\,updates}\right]$
\end{algorithmic}
\end{algorithm}
}

\noindent The brain has the unique capability to adapt to changing environmental conditions through online learning with plastic synapses~\cite{BrainPlasticity} and is able to perform intelligent tasks unattainable yet by computers, while consuming very low  power~\cite{CMead}. Brain-inspired concepts in machine learning applications have so far primarily focused on incorporating the layered, highly interconnected topology of biological neural networks into so-called artificial neural networks (ANNs), including their recurrent variants. RNNs utilizing more complex LSTM or GRU units have demonstrated astounding successes in applications requiring learning from temporal data such as speech recognition or language 
modelling~\cite{Graves2013May, He2019May, LSTMLanguageModelling}.
Recent works have transferred the rich dynamics of the biologically-inspired SNNs to RNNs and demonstrated on-par performance in certain applications by utilizing gradient-based deep learning methods for training~\cite{Bellec2018b, Wozniak2020Jun}. Typically, the aforementioned models are trained with the ubiquitous BPTT algorithm~\cite{Werbos1990Oct}.
Despite the successes of BPTT-trained networks, this algorithm has severe limitations, especially in scenarios involving online learning from a continuous stream of input data. This is because BPTT has to track all past activities by unrolling the network in time, which becomes very deep as the input-sequence length increases. 
For example, a two-second-long spoken input sequence with 1\,ms time steps results in a 2000-layer-deep unrolled network.

Truncated BPTT (TBPTT) reduces the amount of information that needs to be stored by splitting the input stream into shorter segments and by updating the parameters of the network after each segment~\cite{Williams1995Jan}. However, it still leads to so-called system-locking problems~\cite{Jaderberg2016Aug}, because the parameter updates can only be computed after the last input of each segment has been processed. In addition, TBPTT encounters learning deficiencies in tasks where there are long time lags between the network activity and the feedback from the environment.

To address these issues, online learning algorithms were developed for calculating the parameters' updates in real time as the input data arrives. In this manner, they are conceptually more similar to the way the brain adapts to changing environmental conditions. Two such online algorithms were initially introduced in~\cite{Williams1989Jun}. The first one, real-time recurrent learning (RTRL), applies updates to the parameters immediately at each time step, hence sacrificing gradient-equivalence to BPTT, whereas the second one, which we refer to as RTRL with deferred updates, applies BPTT-equivalent updates at the end of the input sequence. However, both versions of RTRL have higher time complexity than BPTT and for this reason remained rarely used in practice.

Recently, online learning algorithms regained popularity and were investigated following two distinct research directions. The first direction focuses on approximations of RTRL with reduced computational complexity. It led to the development of various algorithms, for example Unbiased Online Recurrent Optimization (UORO)~\cite{tallec2018unbiased}, Kronecker Factored RTRL (KF-RTRL)~\cite{Mujika2018} or Optimal Kronecker-Sum Approximation of RTRL (OK-RTRL)~\cite{BenzingGMMS19}, which all provide approximate gradients.
The second direction takes inspiration from biological learning systems and is centered around the debate whether and how they employ error backpropagation~\cite{Lillicrap2019Apr}. Specifically, learning algorithms such as SuperSpike~\cite{Zenke2018May}, Random Feedback Local Online (RFLO)~\cite{RFLO} and e-prop~\cite{Bellec2020Apr} were proposed to approximate the gradients under biological constraints, see~\cite{Marschall2019Jul} for a summary. Note that these approaches were primarily derived for a single large layer of recurrent units and provide approximate gradients.

In this work, we revisit the formulation of gradient-based training and propose a novel online learning methodology that clearly separates the gradient computation into two components: spatial and temporal. Because this gradient separation plays a key role, we refer to our algorithm as Online Spatio-Temporal Learning (OSTL). We derive OSTL for a multi-layer recurrent network of spiking neurons and prove that for the special case of a network with a single recurrent SNN layer the proposed gradient separation maintains equivalence to BPTT. We show that the common feed-forward deep SNN architectures can be trained online with low time complexity and comparable performance to BPTT. Moreover, we extend OSTL for training generic RNNs, such as LSTM networks, with a special focus on multi-layered networks. Our work complements the approximate RTRL-based approaches and the biologically-plausible learning methods by proposing a unified approach that introduces biological insights.
The key contributions of this paper are:

\begin{itemize}
    \item a novel online learning algorithm for multi-layer feed-forward SNNs with a clear separation of spatial and temporal gradients and a time complexity of $O(kn^2)$, where $k$ is the number of layers and $n$ is the largest layer size,
    \item a novel online learning algorithm for shallow feed-forward SNNs with a clear separation of temporal and spatial gradients and a time complexity of $O(n^2)$, while maintaining the gradient equivalence with BPTT,
    \item a generalization of this learning algorithm to generic RNNs with a clear separation of temporal and spatial gradients and a time complexity of $O(kn^4)$,
    \item an explicit derivation of this algorithm for deep recurrent networks comprising spiking neurons, LSTMs and GRUs.
\end{itemize}

\section{Spiking Neural Networks}
\label{sec:SpikingNeuralNetworks}
Inspired by insights from neuroscience, SNNs are network architectures with biologically-realistic neuronal dynamics and synaptic learning mechanisms. While the neuronal dynamics have been successfully abstracted to several neuron models, such as the well-known leaky integrate-and-fire (LIF) neuron model, the training was historically performed with variants of the spike-timing-dependent plasticity (STDP) Hebian rule~\cite{Gerstner2018}. Although STDP is a simple biologically-inspired online learning mechanism, the accuracy of STDP-based architectures is inferior to that of state-of-the-art deep ANNs trained with BPTT. To address this problem, several research groups have focused their activities on facilitating gradient-based training for SNNs~\cite{Bellec2018b, Wozniak2020Jun}. In particular, in~\cite{Wozniak2020Jun} an alternative viewpoint on the spiking neuron was presented, which incorporates the neural dynamics into a recurrent ANN unit called a Spiking Neural Unit (SNU). Thus, in essence, the SNU bridges the ANN world with the SNN world by recasting the SNN dynamics with ANN-based building blocks, see Figure~\ref{fig:SNU}.

\figTwo{t}

The state and output equations for a recurrent SNN layer $l$ composed of $n_l$ SNUs are:
\begin{align}
\boldsymbol{s}_{l}^t &= \mathbf{g}(\mathbf{W}_l \boldsymbol{y}_{l-1}^t + \mathbf{H}_l \boldsymbol{y}_{l}^{t-1} + d \cdot \boldsymbol{s}_{l}^{t-1} \odot (\mathbb{1} - \boldsymbol{y}_{l}^{t-1})) \label{eq:SNUState}\\
\boldsymbol{y}_{l}^t &= \mathbf{h}(\boldsymbol{s}_{l}^t + \mathbf{b}_{l}), \label{eq:SNUOutput}
\end{align}
where $\boldsymbol{s}_{l}^t \in \RR^{n_l}$ represents the internal states, i.e. the membrane potentials, $\boldsymbol{y}_{l}^{t-1} \in \RR^{n_l}$ denotes the output of the neurons of the $l^{th}$ layer at time $t-1$, $\boldsymbol{y}_{l-1}^t \in \RR^{n_{l-1}}$ denotes the output of the neurons of the $(l-1)^{th}$ layer at time $t$, $\mathbf{W}_l \in \RR^{n_l\,\times\,n_{l-1}}$ denotes the input weights from layer $l-1$ to layer $l$, $\mathbf{H}_l \in \RR^{n_l\,\times\,n_l}$ denotes the recurrent weights of layer $l$, $\mathbf{b}_{l} \in \RR^{n_l}$ represents the firing thresholds, $d \in \RR$ is a constant that represents the decay of the membrane potential, and $\mathbf{g}$ and $\mathbf{h}$ are the input and output activation functions, respectively. In order to accurately represent the dynamics of the LIF neuron model, we set $\mathbf{g}=\mathbb{1}$ and $\mathbf{h}=\boldsymbol{\Theta}$, i.e., the Identity and Heaviside functions, respectively. Note that the SNU can also be configured to provide a continuous output, forming a so-called soft SNU (sSNU), for example by using the sigmoid function $\mathbf{h}=\boldsymbol{\sigma}$, and thus operating similarly to other RNN units.
The SNU formulation enables training of multi-layered SNNs through the application of the BPTT approach as illustrated in Figure~\ref{fig:MultiLayerGradientFlowRNN}a.

\section{Online Spatio-Temporal Learning}
\label{sec:OSTL}
Biological neurons maintain a trace of past events in so-called eligibility traces that along with pre- and postsynaptic activities modulate the synaptic plasticity and therefore provide means for temporal credit assignment~\cite{Gerstner2018}. In addition, there exist several so-called learning signals, e.g. dopamine, that transport information spatially from the environment to the individual neurons. We take inspiration from the existence of these two types of signals and introduce OSTL, a novel algorithm for online learning that is addressing the limitations of BPTT discussed so far. To that end, we separate the gradient computation into temporal components corresponding to the eligibility traces and into spatial components corresponding to the learning signals. 

A common objective of learning in neural networks is to train the parameters $\boldsymbol{\theta}$ of the network so that the error $E$ is minimized. In deep SNNs, the network error $E^t \in \RR$ at time $t$ is only a function of the output of the neurons in the last layer $k$ and the target outputs, i.e. $E^t = \phi(\boldsymbol{y}_k^t, \hat{\boldsymbol{y}}^t)$. As highlighted in Section~\ref{sec:SpikingNeuralNetworks}, a generic layer of spiking neurons produces outputs
$\boldsymbol{y}_l^t = \boldsymbol{\chi}(\boldsymbol{s}_l^t, \boldsymbol{\theta}_l)$ and $\boldsymbol{s}_l^t = \boldsymbol{\psi}(\boldsymbol{s}_l^{t-1}, \boldsymbol{y}_l^{t-1}, \boldsymbol{y}_{l-1}^{t}, \boldsymbol{\theta}_l)$, where all the trainable parameters, $\mathbf{W}_l, \mathbf{H}_l, \mathbf{b}_l$, are collectively described by a variable $\boldsymbol{\theta}_l$.


Using this notation, we compute the updates of the parameters $\Delta \boldsymbol{\theta}_l$ that minimize $E$ based on the principle of gradient descent as
\begin{align}
\Delta \boldsymbol{\theta}_l = -\eta \dv{E}{\boldsymbol{\theta}_l}\label{eq:GenericParameterUpdate},
\end{align}
where $\eta \in \RR$ is the learning rate. From Equation~\ref{eq:GenericParameterUpdate}, using the chain rule, the core ingredient of BPTT, we arrive at
\begin{align}
\dv{E}{\boldsymbol{\theta}_l} &= \sum_{1 \leq t \leq T} \pdv{E^t}{\boldsymbol{y}_k^t} \left[ \pdv{\boldsymbol{y}_k^t}{\boldsymbol{s}_k^t} \dv{\boldsymbol{s}_k^t}{\boldsymbol{\theta}_l} + \pdv{\boldsymbol{y}_k^t}{\boldsymbol{\theta}_l} \right], \label{eq:nonexpBPTT}
\end{align}
where the summation ranges from the first time step $t=1$ until the last time step $t=T$. We further expand Equation~\ref{eq:nonexpBPTT} below, in particular $\dv{\boldsymbol{s}_k^t}{\boldsymbol{\theta}_l}$, and isolate the temporal components involving only layer $l$. This allows to unravel a recursion that can be exploited to form an online reformulation of BPTT, see Supplementary material S1. In particular, it can be shown that
\begin{align}
\dv{\boldsymbol{s}_l^t}{\boldsymbol{\theta}_l} &= \sum_{1 \leq \hat{t} \leq t} \left( \prod_{t \ge t' > \hat{t}} \dv{\boldsymbol{s}_l^{t'}}{\boldsymbol{s}_l^{t' - 1}} \right) \left( \pdv{\boldsymbol{s}_l^{\hat{t}}}{\boldsymbol{\theta}_l} + \pdv{\boldsymbol{s}_l^{\hat{t}}}{\boldsymbol{y}_l^{\hat{t}-1}} \pdv{\boldsymbol{y}_l^{\hat{t}-1}}{\boldsymbol{\theta}_l}\right). \label{eq:expandedBPTT}
\end{align}
Equation~\ref{eq:expandedBPTT} can be rewritten in a recursive form as follows
\begin{align}
\boldsymbol{\epsilon}_l^{t,\boldsymbol{\theta}} &\coloneqq \dv{\boldsymbol{s}_l^t}{\boldsymbol{\theta}_l} = \left(\dv{\boldsymbol{s}_l^{t}}{\boldsymbol{s}_l^{t-1}} \boldsymbol{\epsilon}_l^{t-1,\boldsymbol{\theta}} + \left( \pdv{\boldsymbol{s}_l^t}{\boldsymbol{\theta}_l} + \pdv{\boldsymbol{s}_l^t}{\boldsymbol{y}_l^{t-1}} \pdv{\boldsymbol{y}_l^{t-1}}{\boldsymbol{\theta}_l}\right)\right) \label{eq:generalEligibilityVector},
\end{align}
see Supplementary material S1 for a detailed proof. This leads to an expression of the gradient as
\begin{align}
\dv{E}{\boldsymbol{\theta}_l} &=\sum_t \mathbf{L}_l^t \mathbf{e}_l^{t,\boldsymbol{\theta}} + \mathbf{R}\label{eq:OSTLR},
\end{align}
where
\begin{align}
\mathbf{e}^{t,\boldsymbol{\theta}}_{l} &= \pdv{\boldsymbol{y}^t_{l}}{\boldsymbol{s}^t_{l}} \boldsymbol{\epsilon}^{t,\boldsymbol{\theta}}_{l} + \pdv{\boldsymbol{y}^t_{l}}{\boldsymbol{\theta}_{l}} \label{eq:generalEligibilityTrace}\\
\mathbf{L}^t_{l} &= \pdv{E^t}{\boldsymbol{y}_k^t} \left( \prod_{(k-l+1) > m \ge 1} \pdv{\boldsymbol{y}^t_{k-m+1}}{\boldsymbol{s}^t_{k-m+1}} \pdv{\boldsymbol{s}^t_{k-m+1}}{\boldsymbol{y}^t_{k-m}} \right) \label{eq:generalLearningSignal}.
\end{align}
The learning signal can also be recursively expressed as
\begin{align}
\mathbf{L}^t_{l}&= \mathbf{L}^t_{l+1} \left(\pdv{\boldsymbol{y}^t_{l+1}}{\boldsymbol{s}^t_{l+1}} \pdv{\boldsymbol{s}^t_{l+1}}{\boldsymbol{y}^t_{l}}\right)\label{eq:generalLearningSignalRecursive} \,\, \mathrm{with} \,\, \mathbf{L}^t_{k} = \pdv{E^t}{\boldsymbol{y}_k^t}.
\end{align}
Note that in order to simplify the notation, we defined $\boldsymbol{\epsilon}^{t,\boldsymbol{\theta}_l}_{l} \coloneqq \boldsymbol{\epsilon}^{t,\boldsymbol{\theta}}_{l}$ and $\mathbf{e}^{t,\boldsymbol{\theta}_l}_{l} \coloneqq \mathbf{e}^{t,\boldsymbol{\theta}}_{l}$.

The formulation described by Equation~\ref{eq:OSTLR} is novel and constitutes the basis of our online learning scheme, where the emphasis is on separating temporal and spatial gradients. Thus, $\dv{E}{\boldsymbol{\theta}_l}$ is in principle calculated as the sum of individual products of the eligibility traces, $\boldsymbol{e}^{t,\mathbf{W}_l}_{l}, \boldsymbol{e}^{t,\mathbf{H}_l}_{l}, \boldsymbol{e}^{t,\mathbf{b}_l}_{l}$, which are collectively described by $\boldsymbol{e}^{t,\boldsymbol{\theta}_l}_{l}$
defined in Equation~\ref{eq:generalEligibilityTrace}, and the learning signal $\mathbf{L}^t_{l} \in \RR^{n_l}$, defined in Equation~\ref{eq:generalLearningSignal}. 
Following the biological inspiration, the eligibility trace contains only the temporal component of the gradient computation. It evolves 
through the recursive Equation~\ref{eq:generalEligibilityVector} 
that involves only information local to a layer of spiking neurons. The learning signal is applied locally in time and corresponds to the spatial gradient component that evolves through the recursive Equation~\ref{eq:generalLearningSignalRecursive}.
Finally, the residual term $\mathbf{R}$ represents the combinations of spatial and temporal gradients. Specifically, this term appears because the total derivative $\dv{\boldsymbol{s}_k^t}{\boldsymbol{\theta}_l}$ in Equation~\ref{eq:nonexpBPTT} involves cross-layers dependencies, see Supplementary material S1 for more details.

\subsection{OSTL for multi-layer SNNs}

\figThree{!ht}

The main aim of OSTL is to maintain the separation between spatial and temporal gradients. For deep networks we explore an approach in which we simplify Equation~\ref{eq:OSTLR} by omitting the term $\mathbf{R}$. This approximation leads to the following formulation of OSTL for multi-layer SNNs
\begin{align}
\dv{E}{\boldsymbol{\theta}_{l}} &= \sum_t \mathbf{L}^t_{l} \mathbf{e}^{t,\boldsymbol{\theta}}_{l}.\label{eq:OSTL}
\end{align}
Therefore, the gradient at each time step $\dv{E^t}{\boldsymbol{\theta}_l} = \mathbf{L}^t_{l} \mathbf{e}^{t,\boldsymbol{\theta}}_{l}$ and eventually the parameter updates $\Delta \boldsymbol{\theta}_l$ can be computed online using the eligibility traces and the learning signals, as shown in Figure~\ref{fig:MultiLayerGradientFlowRNN}b.

Below we provide the explicit formulation of OSTL for the commonly used multi-layered feed-forward SNU network. 
In this case, the recurrency matrix $\mathbf{H}$ is absent and Equation~\ref{eq:generalEligibilityTrace} becomes
\begin{align}
\mathbf{e}_{l}^{t,\mathbf{W}} &= \mathrm{diag}\left({\mathbf{h'}}_l^t\right) \boldsymbol{\epsilon}_{l}^{t,\mathbf{W}}\label{eq:FFExplicitEW}\\
\mathbf{e}_{l}^{t,\mathbf{b}} &= \mathrm{diag}\left({\mathbf{h'}}_l^t\right) \boldsymbol{\epsilon}_{l}^{t,\mathbf{b}} + \mathrm{diag}\left({\mathbf{h'}}_l^t\right)\label{eq:FFExplicitEb},
\end{align}
where
\begin{align}
\boldsymbol{\epsilon}_{l}^{t,\mathbf{W}} &= \dv{\boldsymbol{s}_{l}^t}{\boldsymbol{s}_{l}^{t-1}} \boldsymbol{\epsilon}_{l}^{t-1,\mathbf{W}} + \mathrm{diag}\left({\mathbf{g'}}_l^t\right) \mathrm{diag}\left(\boldsymbol{\Upsilon}^t_{l-1}\right)\label{eq:FFExplicitEpsW}\\
\boldsymbol{\epsilon}_{l}^{t,\mathbf{b}} &= \dv{\boldsymbol{s}_{l}^t}{\boldsymbol{s}_{l}^{t-1}} \boldsymbol{\epsilon}_{l}^{t-1,\mathbf{b}} - d \cdot \mathrm{diag}\left({\mathbf{g'}}_l^t\right) \mathrm{diag}\left(\boldsymbol{s}_{l}^{t-1} \odot {\mathbf{h'}}_l^{t-1} \right)\label{eq:FFExplicitEpsb}
\end{align}
and
\begin{align}
\dv{\boldsymbol{s}_{l}^t}{\boldsymbol{s}_{l}^{t-1}} = \mathrm{diag}\left({\mathbf{g'}}_l^t\right) & d \cdot \mathrm{diag} \left( \left( \mathbb{1} - \boldsymbol{y}_{l}^{t-1}\right) -\boldsymbol{s}_{l}^{t-1} \odot {\mathbf{h'}}_l^{t-1}\right)\label{eq:DeepSNNdsds}.
\end{align}
Note that we have used the short-hand notation of $\dv{\mathbf{g}(\boldsymbol{\xi}_l^t)}{\boldsymbol{\xi}_l^t} = {\mathbf{g'}}_l^t$, $\dv{\mathbf{h}(\boldsymbol{\xi}_l^t)}{\boldsymbol{\xi}_l^t} = {\mathbf{h'}}_l^t$. Furthermore, the elements of $\boldsymbol{\Upsilon}^t_{l}$ are given by $\left(\Upsilon^t_{l}\right)_{opq} = \delta_{op} \left(y_{l}^t\right)_q$.

For a mean squared error loss with readout weights, 
\begin{align}
E^t = \frac{1}{2} \sum_{n_k} \left(\mathbf{W}^{\text{out}} \boldsymbol{y}_k^t - \hat{\boldsymbol{y}}^t \right)^2\label{eq:MeanSquaredLossFunction},
\end{align}
where 
$\mathbf{W}^{\text{out}} \in \RR^{n_k\,\mathrm{x}\,n_k}$ are the readout weights, the learning signal stated in Equation~\ref{eq:generalLearningSignal} is
\begin{align}
\mathbf{L}_{l}^t = \mathbf{L}_{l+1}^t \left(\mathrm{diag}\left(\mathbf{h'}_{l+1}^t \odot \mathbf{g'}_{l+1}^t\right) \mathbf{W}_{l+1} \right)\label{eq:FFExplicitLearningSignal},
\end{align}
with
\begin{align}
\mathbf{L}_{k}^t = \left(\left(\mathbf{W}^{\text{out}}\right)^T \boldsymbol{y}_k^t - \hat{\boldsymbol{y}}^t\right).
\end{align}

Algorithm~\ref{alg:OSTL} illustrates in a pseudocode the steps necessary to compute the parameter updates which may be applied immediately or deferred, similarly to RTRL. Note that, we further focus on the latter case. The particular equations in Algorithm~\ref{alg:OSTL} depend on the units and the network architecture. Although the SNN described by Equations~\ref{eq:FFExplicitEW}-\ref{eq:DeepSNNdsds} is feed-forward, this does not prevent solving temporal tasks. In such architectures, including also quasi-recurrent neural networks~\cite{Bradbury2016Nov}, the network relies on the internal states of the units rather than on layer-wise recurrency matrices $\mathbf{H}_l$. In addition, such architectures allow to greatly reduce the time complexity of OSTL to $O(k n^2)$, compared to $O(k n^4)$ for recurrent multi-layered SNNs. In the latter case, the time complexity is primarily dominated by the recurrency matrix $\mathbf{H}_{l}$, see Supplementary material S2.
By neglecting the influence of the recurrent matrix $\mathbf{H}$ for the eligibility traces (OSTL w/o $\mathbf{H}$), the complexity of recurrent multi-layered SNN can also be reduced to $O(k n^2)$, see Supplementary material S3.

\algoOne{!t}

\subsection{OSTL for single-layer SNNs}
In case of a single recurrent layer, there is no residual term $\mathbf{R}$ emerging in Equation~\ref{eq:OSTLR}, that is $\mathbf{R}=0$, see Supplementary material S1. For this architecture, the additional equations needed besides Equations~\ref{eq:FFExplicitEW}-\ref{eq:FFExplicitEpsW} are
\begin{align}
\mathbf{e}^{t,\mathbf{H}} &= \mathrm{diag}(\mathbf{h'})^t \boldsymbol{\epsilon}^{t,\mathbf{H}}\\
\boldsymbol{\epsilon}^{t,\mathbf{H}} &=  \dv{\boldsymbol{s}^t}{\boldsymbol{s}^{t-1}} \boldsymbol{\epsilon}^{t-1,\mathbf{H}} +
\mathrm{diag}\left({\mathbf{g'}}^t\right) \mathrm{diag}\left(\boldsymbol{\Upsilon}^{t-1}\right)
\end{align}
\begin{align}
\boldsymbol{\epsilon}^{t,\mathbf{b}} &= \dv{\boldsymbol{s}^t}{\boldsymbol{s}^{t-1}} \boldsymbol{\epsilon}^{t-1,\mathbf{b}} + \mathbf{H} \mathrm{diag}\left({\mathbf{h'}}^{t-1}\right) \nonumber\\
&- d \cdot \mathrm{diag}\left({\mathbf{g'}}^t\right) \mathrm{diag}\left(\boldsymbol{s}^{t-1} \odot {\mathbf{h'}}^{t-1} \right)\label{eq:RecExplicitEb}
\end{align}
and
\begin{align}
\dv{\boldsymbol{s}^{t}}{\boldsymbol{s}^{t-1}} &= \mathrm{diag}\left(\mathbf{g'}^t\right) \left[\mathbf{H} \mathrm{diag}\left({\mathbf{h'}}^{t-1}\right) \right.\nonumber\\
&+ \left. d \cdot \mathrm{diag}\left( \left( \mathbb{1} - \boldsymbol{y}^{t-1}\right) - \boldsymbol{s}^{t-1} \odot {\mathbf{h'}}^{t-1}\right) \right]. \label{eq:totalDerivativeS}
\end{align}
Using the loss from Equation~\ref{eq:MeanSquaredLossFunction}, the learning signal is
\begin{align}
\mathbf{L}^t &= \pdv{E^t}{\boldsymbol{y}^t} = \left(\mathbf{W}^{\text{out}}\right)^T \left(\boldsymbol{y}^t - \hat{\boldsymbol{y}}^t\right)\label{eq:LearningSignalSingleLayer}.
\end{align}
Note that in this special case of a single layer network the subscript $l$, signifying the layer, has been removed.
Because $\mathbf{R}=0$, there are no approximations involved and OSTL is gradient-equivalent to BPTT. See also the formal proof in Supplementary material S1. Moreover, the gradient-equivalence is maintained even if this recurrent layer is embedded in a deep network architecture comprising an arbitrary number of stateless layers, e.g. a series of convolutions followed by the recurrent layer and a softmax output layer, see Supplementary material S4. This important property has also been empirically demonstrated via simulations.

Note that while the eligibility traces are biologically inspired and involve only information local to a layer, the learning signal relies on the mathematical concept of error backpropagation applied locally in time. However, more stringent biological constraints may be introduced into OSTL.
For example, the requirement of error backpropagation in Equation~\ref{eq:generalLearningSignal} can be relaxed with biologically plausible credit assignment schemes, such as random feedback alignment. 
See Supplementary material S3 for the details.

\tabTwo{!ht}

\subsection{Generalization of OSTL to generic RNNs}
\label{sec:OSTLforLSTM}
In the sequel, we extend OSTL to generic deep recurrent neural networks, for example consisting of LSTM units. We note that such units obey more complex state and output equations compared to
SNNs. In particular, the output $\boldsymbol{y}_l^t$ is a recursive function and depends on the outputs of the previous layer as well as the trainable parameters, that is:
\begin{align}
\boldsymbol{y}_l^t &= \boldsymbol{\chi}(\boldsymbol{s}_l^t, \boldsymbol{y}_l^{t-1}, \boldsymbol{y}_{l-1}^{t}, \boldsymbol{\theta}_l),\label{eq:multilayerOutput}\\
\boldsymbol{s}_l^t &= \boldsymbol{\psi}(\boldsymbol{s}_l^{t-1}, \boldsymbol{y}_l^{t-1}, \boldsymbol{y}_{l-1}^{t}, \boldsymbol{\theta}_l).\label{eq:multilayerState}
\end{align}
To the best of our knowledge, this generalized form covers all standard RNN units currently used in literature.

Performing similar steps as in the previous subsections, we derive the eligibility trace for layer $l$ as
\begin{align}
\mathbf{e}^{t,\boldsymbol{\theta}}_{l} &= \pdv{\boldsymbol{y}^t_{l}}{\boldsymbol{s}^t_{l}} \boldsymbol{\epsilon}_{l}^{t,\boldsymbol{\theta}} + \pdv{\boldsymbol{y}^t_{l}}{\boldsymbol{\theta}_{l}} + \pdv{\boldsymbol{y}^t_{l}}{\boldsymbol{y}^{t-1}_{l}} \mathbf{e}^{t-1,\boldsymbol{\theta}}_{l}.\label{eq:generalEligibilityTraceExtended}
\end{align}
Note that compared to the case of SNNs, the eligibility trace contains an additional recurrent term, which is the eligibility trace of the previous time step $t-1$, i.e. $\mathbf{e}_{l}^{t-1,\boldsymbol{\theta}}$.
In consequence, these formulations expand the concept of the eligibility traces of a particular neuron to depend also on activities of other neurons within the same layer. Nevertheless, the core function of the eligibility traces, to capture the temporal gradient contributions, is maintained.

The generalized learning signal is
\begin{align}
\mathbf{L}^t_{l} = \mathbf{L}^t_{l+1} \left(\left[\pdv{\boldsymbol{y}^t_{l+1}}{\boldsymbol{s}^t_{l+1}} \pdv{\boldsymbol{s}^t_{l+1}}{\boldsymbol{y}^t_{l}} + \pdv{\boldsymbol{y}^t_{l+1}}{\boldsymbol{y}^t_{l}}\right] \right)  \label{eq:generalLearningSignalExtended} \,\, \mathrm{with} \,\, \mathbf{L}^t_{k} = \pdv{E^t}{\boldsymbol{y}_k^t}.
\end{align}
Finally, similarly to the case of deep SNN, the parameter update $\dv{E}{\boldsymbol{\theta}_l}$ is computed using Equation~\ref{eq:OSTL}. Therefore, the OSTL extension described in Equations~\ref{eq:generalEligibilityTraceExtended}-\ref{eq:generalLearningSignalExtended} enables online training for a broad variety of RNN units, including LSTMs and GRUs, for which the
explicit formulas can be found in Supplementary material S5 and S6, respectively. 

\section{Comparison to related algorithms}
Table~\ref{complexityComparisonTable} summarizes the properties of various on-line learning algorithms including OSTL. The horizontal lines define three groups: the fundamental algorithms, i.e. BPTT and RTRL, the recently introduced approximate algorithms, and various OSTL versions. However, two algorithms from the prior art, namely RTRL and e-prop, share important concepts with OSTL and therefore deserve further discussion. 

OSTL and RTRL are derived by optimizing the same objective function using the gradient descent principles, but the explicit treatment of the gradient separation in OSTL leads to a different factorization of the operations that rely on layer-local rather than global information. In other words, the eligibility traces and learning signals of OSTL are defined on a per layer basis rather than on the complete network basis. Thus, for multi-layered networks, OSTL scales with the size of the largest layer, whereas RTRL with the total number of neurons of a global pool of neurons.

Although the generic form of Equation~\ref{eq:OSTL} appears to be similar to e-prop, there are several critical differences. Firstly, the eligibility traces and learning signals of OSTL described in Equations~\ref{eq:generalEligibilityTrace} and~\ref{eq:generalLearningSignal} have been derived without approximations and are different than the corresponding ones in e-prop given in Equations~4 and~13 presented in~\cite{Bellec2020Apr}. Specifically, OSTL uses the total derivative $\dv{\boldsymbol{s}_l^{t}}{\boldsymbol{s}_l^{t-1}}$, while e-prop uses the partial derivative $\pdv{\boldsymbol{s}_l^{t}}{\boldsymbol{s}_l^{t-1}}$.

This subtle, yet crucial difference leads to different learning results. For example, in a recurrently connected layer of SNUs described by Equations~\ref{eq:SNUState} and~\ref{eq:SNUOutput}, the differences between the total derivative in Equation~\ref{eq:totalDerivativeS} and the partial derivative given below
\begin{align}
\pdv{\boldsymbol{s}_l^{t}}{\boldsymbol{s}_l^{t-1}} &= \mathrm{diag}\left(\mathbf{g'}_l^t\right) \left[d \cdot \mathrm{diag}\left(\mathbb{1} - \boldsymbol{y}_{l}^{t-1}\right)\right]
\end{align}
become apparent, as the terms involving $\dv{\boldsymbol{y}_{l}^{t}(\boldsymbol{s}_{l}^{t})}{\boldsymbol{s}_{l}^{t}}$ are missing.

The definition of the learning signal is also different. OSTL builds on a clear separation of spatial and temporal gradients without any approximations. Therefore, the learning signal $\mathbf{L}_l^t$ of OSTL, stated in Equation~\ref{eq:LearningSignalSingleLayer}, is defined as the exact gradient signal from the output layer w.r.t all layers of the network. 
In contrast, e-prop mostly focuses on single-layer networks and defines the learning signal, following the general form given in Equation~4 of~\cite{Bellec2020Apr}
\begin{align}
\mathbf{L}_{\text{e-prop}}^t = \mathbf{B} (\boldsymbol{y}^t - \boldsymbol{y}^{*,t}),
\end{align}
where $\mathbf{B}$, depending on the variant of e-prop, is either the matrix of output weights, or a random weight matrix. The difference of the learning signal is especially predominant in the case of a multi-layered network, where in e-prop every neuron in each layer is required to have a direct connection to the output layer, while the learning signal of OSTL is defined for every neuron, even in layers without a direct connection to the output layer.

\figJoint{!ht}

\section{Results}
\label{sec:Results}
We evaluated the performance of OSTL on four tasks: music prediction, handwritten digit classification, language modelling and speech recognition. For each task, we compared the accuracy of OSTL with that of BPTT-based training. The setup details are described in Supplementary material S7.

Firstly, for the music prediction task based on the JSB dataset~\cite{JSB}, we employed an architecture comprising a single RNN layer with 150 units and a stateless output layer comprising 88 sigmoid units, as in \cite{Wozniak2020Jun}. In particular, we analyzed three types of RNN layers, an SNU layer, an sSNU layer as well as an LSTM layer. For the SNU architecture, OSTL provides a low-complexity formulation, i.e., $O(n^2)$, with BPTT-equivalent gradients. In order to demonstrate this aspect, the same networks, including all its hyperparameters, were trained with OSTL and BPTT.
As shown in Figure~\ref{fig:Results}a, the results obtained with OSTL are on par with these obtained with BPTT, thereby empirically validating the gradient equivalence of BPTT and OSTL. In addition, we trained the SNU network with OSTL using random feedback alignment (OSTL rnd) and random e-prop. Both, for OSTL with random feedback and for e-prop, the neg. log-likelihood increases by $0.16$, which we attribute to the random feedback matrix.

The second task illustrates the low-complexity formulation of OSTL for deep feed-forward SNNs using the handwritten digit recognition MNIST dataset~\cite{Lecun1998Nov}. In particular, we employed two hidden layers of 256 units each followed by 10 output units. Note that all units were of SNU or sSNU type. Following the standard SNN convention, we used biologically-inspired rate coding that introduces a temporal aspect to this dataset by coding the grey scale values of the pixels in the rates of input spikes
~\cite{GerstnerRateCoding}. As discussed in Section~\ref{sec:OSTL}, it is possible to train such a network online at time complexity of $O(k n^2)$. Although OSTL for deep networks omits the residual term $\mathbf{R}$, it achieved competitive classification accuracies to the BPTT baseline as illustrated in Figure~\ref{fig:Results}b.

As a third task, we applied OSTL to language modelling based on the Penn Tree Bank (PTB) dataset~\cite{PTB}. To that end, we trained an architecture comprising a fully-connected layer of 10000 units for embedding, three layers of 1300 feed-forward sSNUs and a softmax output layer of 10000 units. Also in this difficult task, OSTL achieved very competitive performance compared to TBPTT with a truncation horizon of $T=20$ words, see Figure~\ref{fig:Results}c.

Lastly, we considered an application to speech recognition based on the TIMIT dataset~\cite{TIMIT}. This task requires to capture long temporal dependencies between the inputs and the outputs, which is especially challenging for TBPTT. We analyze this by varying the truncation horizon $T \in \{10,30,50,90\}$. The performance of TBPTT is compared to OSTL and BPTT with complete unrolling. For this task we employed a network with 39 input neurons, two layers of 200 recurrent sSNUs or LSTM units and a softmax output layer of 39 units. The results in Figure~\ref{fig:Results}d show the percentage error rate. The decreasing performance of TBPTT with shorter truncation horizons $T$ indicates that the substantial overhead of unrolling the network far into the past is essential.
OSTL in contrast, despite the lack of full BPTT-equivalence,
demonstrates competitive performance without unrolling. 
In addition, we trained a reduced network with one hidden layer with OSTL w/o $\mathbf{H}$, where the influence of the recurrent matrix $\mathbf{H}$ for the eligibility traces is neglected. Despite the slight performance degradation of OSTL w/o $\mathbf{H}$ of $4\%$, the time complexity is reduced drastically to $O(k n^2)$. For comparison, we trained this network with random e-prop, and observe a more significant performance drop of $15\%$.


\section{Conclusion}
OSTL is a novel online learning algorithm, whose basic form provides gradients equivalent to those of BPTT and RTRL with deferred updates. However, in contrast to these algorithms, it specifically focuses on the biologically-inspired separation of spatial and temporal gradient components, which facilitates studies of credit assignment schemes and further development of approximate online learning algorithms.
In particular, OSTL lends itself to efficient online training of $k$-layer feed-forward SNU-based networks with a time complexity of $O(k n^2)$, where the update of the synaptic weights is decomposed into a learning signal and an eligibility trace as in the three-factor learning rules. Even more so, it enables online training of shallow feed-forward SNNs with BPTT-equivalent gradients and time complexity of $O(n^2)$. Moreover, in the case of recurrent single-layer SNNs by omitting the influence of the $\mathbf{H}$ matrix in the eligibility traces the complexity remains at $O(n^2)$, without substantial loss in performance.  Finally, OSTL has been further generalized to deep RNNs comprising spiking neurons or more complex units, such as LSTMs and GRUs, demonstrating competitive accuracy in comparison to BPTT. The proposed algorithm allows efficient online training for temporal data and opens a new avenue for the adoption of trainable recurrent networks in low-power IoT and edge AI devices.

\section*{Broader Impact}
Online learning for temporal data provides a viable alternative to the traditional training with backpropagation-through-time, which is typically  implemented offline in large-scale processing systems. It supports continuous adaptation of the AI system model to changes in the observed environment. This characteristic will enable accurate and low-complexity on-chip learning, thus providing a path for the deployment of training on IoT and edge devices, in particular for supporting applications that the input data patterns are dynamically changing. Besides broadening the AI application space, it has the potential to limit privacy issues. In cases where private data is required for training, online learning can ensure that the data is solely used on the device and not sent to a third party, as it is common in contemporary AI systems. 

Given that this learning methodology allows online changes to the AI model, a potential issue to be considered relates to adversarial data inputs. The effect can be understood considering the example of a teacher giving students wrong information, who tend to believe the supervisor. This might lead to unintended behavior of the AI system that depending on the application may have severe consequences. Therefore, mechanisms should be put in place to monitor and prevent such attempts. Clearly, online learning should be avoided in use cases that either is not required, e.g. for repetitive tasks, or not allowed because of regulatory reasons, e.g. in autonomous driving vehicles. 

Since AI systems become ubiquitous today, the training mechanism of such systems has a great impact. As is outlined here, online learning schemes can be very beneficial when implemented and handled properly, however they may also exhibit drastic negative impact when applied carelessly.

\section{ Acknowledgments}
This project has been funded trough the ERA-NET CHIST-ERA programme by SNSF under the project number 20CH21\_186999 / 1.

\bibliography{bibliography}
\clearpage

\newpage
\newpage


\section*{Supplementary material}

# Supplementary Material for Online Spatio-Temporal Learning in Deep Neural Networks

### Supplementary material S1: Detailed derivation of OSTL for SNNs

This section focuses on the derivation of OSTL for SNNs including the proof of gradient equivalence to BPTT in the case of a single layer.

**Derivation of OSTL for single-layer networks**

As presented in the main text, the state and the output Equations 1 and 2 of a single-layer SNN can be generalized to

$$\boldsymbol{s}^t = \boldsymbol{\psi}(\boldsymbol{y}_0^t, \boldsymbol{s}^{t-1}, \boldsymbol{y}^{t-1}, \boldsymbol{\theta})$$
$$\boldsymbol{y}^t = \boldsymbol{\chi}^t(\boldsymbol{s}^t, \boldsymbol{\theta}).$$

The starting point for the derivation is the calculation of the derivative of the loss function $E$ with respect to the parameters $\boldsymbol{\theta}$. Based on the chain rule, we express $\frac{\mathrm{d}E}{\mathrm{d}\boldsymbol{\theta}}$ as

$$\frac{\mathrm{d}E}{\mathrm{d}\boldsymbol{\theta}} = \sum_{1 \leq t \leq T} \frac{\partial E^t}{\partial \boldsymbol{y}^t} \left[ \frac{\partial \boldsymbol{y}^t}{\partial \boldsymbol{s}^t} \frac{\mathrm{d}\boldsymbol{s}^t}{\mathrm{d}\boldsymbol{\theta}} + \frac{\partial \boldsymbol{y}^t}{\partial \boldsymbol{\theta}} \right]. \tag{1}$$

We will prove that

$$\frac{\mathrm{d}\boldsymbol{s}^t}{\mathrm{d}\boldsymbol{\theta}} = \sum_{1 \leq \hat{t} \leq t} \left( \prod_{t \geq t' > \hat{t}} \frac{\mathrm{d}\boldsymbol{s}^{t'}}{\mathrm{d}\boldsymbol{s}^{t'-1}} \right) \left( \frac{\partial \boldsymbol{s}^{\hat{t}}}{\partial \boldsymbol{\theta}} + \frac{\partial \boldsymbol{s}^{\hat{t}}}{\partial \boldsymbol{y}^{\hat{t}-1}} \frac{\partial \boldsymbol{y}^{\hat{t}-1}}{\partial \boldsymbol{\theta}} \right). \tag{2}$$

To validate the equality in Equation 2 we resort to a proof by induction. The induction basis for the first step of $t = 1$ for the left-hand side becomes

$$\left.\frac{\mathrm{d}\boldsymbol{s}^t}{\mathrm{d}\boldsymbol{\theta}}\right|_{t=1} = \frac{\mathrm{d}\boldsymbol{s}^1}{\mathrm{d}\boldsymbol{\theta}} = \frac{\partial \boldsymbol{s}^1}{\partial \boldsymbol{\theta}} \tag{3}$$

and for the right-hand side

$$\left.\sum_{1 \leq \hat{t} \leq t} \left( \prod_{t \geq t' > \hat{t}} \frac{\mathrm{d}\boldsymbol{s}^{t'}}{\mathrm{d}\boldsymbol{s}^{t'-1}} \right) \left( \frac{\partial \boldsymbol{s}^{\hat{t}}}{\partial \boldsymbol{\theta}} + \frac{\partial \boldsymbol{s}^{\hat{t}}}{\partial \boldsymbol{y}^{\hat{t}-1}} \frac{\partial \boldsymbol{y}^{\hat{t}-1}}{\partial \boldsymbol{\theta}} \right)\right|_{t=1} = \sum_{1 \leq \hat{t} \leq 1} \left( \frac{\partial \boldsymbol{s}^{\hat{t}}}{\partial \boldsymbol{\theta}} + \frac{\partial \boldsymbol{s}^{\hat{t}}}{\partial \boldsymbol{y}^{\hat{t}-1}} \frac{\partial \boldsymbol{y}^{\hat{t}-1}}{\partial \boldsymbol{\theta}} \right) = \frac{\partial \boldsymbol{s}^1}{\partial \boldsymbol{\theta}}. \tag{4}$$

Having verified the induction basis, the induction step needs to be investigated

$$\frac{\mathrm{d}\boldsymbol{s}^{t+1}}{\mathrm{d}\boldsymbol{\theta}} = \frac{\mathrm{d}\boldsymbol{s}^{t+1}}{\mathrm{d}\boldsymbol{s}^t} \frac{\mathrm{d}\boldsymbol{s}^t}{\mathrm{d}\boldsymbol{\theta}} + \frac{\partial \boldsymbol{s}^{t+1}}{\partial \boldsymbol{y}^t} \frac{\partial \boldsymbol{y}^t}{\partial \boldsymbol{\theta}} + \frac{\partial \boldsymbol{s}^{t+1}}{\partial \boldsymbol{\theta}} \tag{5}$$

$$= \left( \frac{\mathrm{d}\boldsymbol{s}^{t+1}}{\mathrm{d}\boldsymbol{s}^t} \right) \left[ \sum_{1 \leq \hat{t} \leq t} \left( \prod_{t \geq t' > \hat{t}} \frac{\mathrm{d}\boldsymbol{s}^{t'}}{\mathrm{d}\boldsymbol{s}^{t'-1}} \right) \left( \frac{\partial \boldsymbol{s}^{\hat{t}}}{\partial \boldsymbol{\theta}} + \frac{\partial \boldsymbol{s}^{\hat{t}}}{\partial \boldsymbol{y}^{\hat{t}-1}} \frac{\partial \boldsymbol{y}^{\hat{t}-1}}{\partial \boldsymbol{\theta}} \right) \right] + \frac{\partial \boldsymbol{s}^{t+1}}{\partial \boldsymbol{y}^t} \frac{\partial \boldsymbol{y}^t}{\partial \boldsymbol{\theta}} + \frac{\partial \boldsymbol{s}^{t+1}}{\partial \boldsymbol{\theta}}$$

$$= \left[ \sum_{1 \leq \hat{t} \leq t} \left( \prod_{(t+1) \geq t' > \hat{t}} \frac{\mathrm{d}\boldsymbol{s}^{t'}}{\mathrm{d}\boldsymbol{s}^{t'-1}} \right) \left( \frac{\partial \boldsymbol{s}^{\hat{t}}}{\partial \boldsymbol{\theta}} + \frac{\partial \boldsymbol{s}^{\hat{t}}}{\partial \boldsymbol{y}^{\hat{t}-1}} \frac{\partial \boldsymbol{y}^{\hat{t}-1}}{\partial \boldsymbol{\theta}} \right) \right] + \frac{\partial \boldsymbol{s}^{t+1}}{\partial \boldsymbol{y}^t} \frac{\partial \boldsymbol{y}^t}{\partial \boldsymbol{\theta}} + \frac{\partial \boldsymbol{s}^{t+1}}{\partial \boldsymbol{\theta}}$$

$$\frac{\mathrm{d}\boldsymbol{s}^{t+1}}{\mathrm{d}\boldsymbol{\theta}} = \sum_{1 \leq \hat{t} \leq (t+1)} \left( \prod_{(t+1) \geq t' > \hat{t}} \frac{\mathrm{d}\boldsymbol{s}^{t'}}{\mathrm{d}\boldsymbol{s}^{t'-1}} \right) \left( \frac{\partial \boldsymbol{s}^{\hat{t}}}{\partial \boldsymbol{\theta}} + \frac{\partial \boldsymbol{s}^{\hat{t}}}{\partial \boldsymbol{y}^{\hat{t}-1}} \frac{\partial \boldsymbol{y}^{\hat{t}-1}}{\partial \boldsymbol{\theta}} \right). \tag{6}$$

By comparing Equation 6 with the hypothesis in Equation 2 for $t+1$, it is evident that this induction step is also fulfilled, which completes the proof.

Therefore, we can expand Equation 1 to

$$\frac{\mathrm{d}E}{\mathrm{d}\boldsymbol{\theta}} = \sum_{1 \leq t \leq T} \frac{\partial E^t}{\partial \boldsymbol{y}^t} \left[ \frac{\partial \boldsymbol{y}^t}{\partial \boldsymbol{s}^t} \sum_{1 \leq \hat{t} \leq t} \left( \prod_{t \geq t' > \hat{t}} \frac{\mathrm{d}\boldsymbol{s}^{t'}}{\mathrm{d}\boldsymbol{s}^{t'-1}} \right) \left( \frac{\partial \boldsymbol{s}^{\hat{t}}}{\partial \boldsymbol{\theta}} + \frac{\partial \boldsymbol{s}^{\hat{t}}}{\partial \boldsymbol{y}^{\hat{t}-1}} \frac{\partial \boldsymbol{y}^{\hat{t}-1}}{\partial \boldsymbol{\theta}} \right) + \frac{\partial \boldsymbol{y}^t}{\partial \boldsymbol{\theta}} \right]. \quad (7)$$

We analyze now the term $\frac{\mathrm{d}\boldsymbol{s}^t}{\mathrm{d}\boldsymbol{\theta}}$ in Equation 2. At time $t+1$, the term $\frac{\mathrm{d}\boldsymbol{s}^{t+1}}{\mathrm{d}\boldsymbol{\theta}}$ can be obtained by multiplying $\frac{\mathrm{d}\boldsymbol{s}^t}{\mathrm{d}\boldsymbol{\theta}}$ by $\frac{\mathrm{d}\boldsymbol{s}^{t+1}}{\mathrm{d}\boldsymbol{s}^t}$ and adding the term $\left( \frac{\partial \boldsymbol{s}^{t+1}}{\partial \boldsymbol{y}^t} \frac{\partial \boldsymbol{y}^t}{\partial \boldsymbol{\theta}} + \frac{\partial \boldsymbol{s}^{t+1}}{\partial \boldsymbol{\theta}} \right)$, thus we obtain Equation 5. This leads to the following recursive reformulation of Equation 2

$$\frac{\mathrm{d}\boldsymbol{s}^t}{\mathrm{d}\boldsymbol{\theta}} = \left( \frac{\mathrm{d}\boldsymbol{s}^t}{\mathrm{d}\boldsymbol{s}^{t-1}} \frac{\mathrm{d}\boldsymbol{s}^{t-1}}{\mathrm{d}\boldsymbol{\theta}} + \left( \frac{\partial \boldsymbol{s}^t}{\partial \boldsymbol{\theta}} + \frac{\partial \boldsymbol{s}^t}{\partial \boldsymbol{y}^{t-1}} \frac{\partial \boldsymbol{y}^{t-1}}{\partial \boldsymbol{\theta}} \right) \right). \quad (8)$$

We define the eligibility vector $\boldsymbol{\epsilon}^{t,\boldsymbol{\theta}}$ to be

$$\boldsymbol{\epsilon}^{t,\boldsymbol{\theta}} := \frac{\mathrm{d}\boldsymbol{s}^t}{\mathrm{d}\boldsymbol{\theta}}, \quad (9)$$

which can also be recursively expressed as,

$$\boldsymbol{\epsilon}^{t,\boldsymbol{\theta}} = \left( \frac{\mathrm{d}\boldsymbol{s}^t}{\mathrm{d}\boldsymbol{s}^{t-1}} \boldsymbol{\epsilon}^{t-1,\boldsymbol{\theta}} + \left( \frac{\partial \boldsymbol{s}^t}{\partial \boldsymbol{\theta}} + \frac{\partial \boldsymbol{s}^t}{\partial \boldsymbol{y}^{t-1}} \frac{\partial \boldsymbol{y}^{t-1}}{\partial \boldsymbol{\theta}} \right) \right). \quad (10)$$

Note that this equation corresponds to Equation 6 of the main text for a single layer. In the case of $t = 1$ the eligibility vector has the special form

$$\boldsymbol{\epsilon}^{1,\boldsymbol{\theta}} = \frac{\mathrm{d}\boldsymbol{s}^1}{\mathrm{d}\boldsymbol{\theta}} = \frac{\partial \boldsymbol{s}^1}{\partial \boldsymbol{\theta}}. \quad (11)$$

Next, we rewrite Equation 7 using Equation 10 as

$$\frac{\mathrm{d}E}{\mathrm{d}\boldsymbol{\theta}} = \sum_{1 \leq t \leq T} \frac{\partial E^t}{\partial \boldsymbol{y}^t} \left[ \frac{\partial \boldsymbol{y}^t}{\partial \boldsymbol{s}^t} \left( \frac{\mathrm{d}\boldsymbol{s}^t}{\mathrm{d}\boldsymbol{s}^{t-1}} \boldsymbol{\epsilon}^{t-1,\boldsymbol{\theta}} + \left( \frac{\partial \boldsymbol{s}^t}{\partial \boldsymbol{\theta}} + \frac{\partial \boldsymbol{s}^t}{\partial \boldsymbol{y}^{t-1}} \frac{\partial \boldsymbol{y}^{t-1}}{\partial \boldsymbol{\theta}} \right) \right) + \frac{\partial \boldsymbol{y}^t}{\partial \boldsymbol{\theta}} \right]$$

$$= \sum_{1 \leq t \leq T} \mathbf{L}^t \left[ \frac{\partial \boldsymbol{y}^t}{\partial \boldsymbol{s}^t} \left( \frac{\mathrm{d}\boldsymbol{s}^t}{\mathrm{d}\boldsymbol{s}^{t-1}} \boldsymbol{\epsilon}^{t-1,\boldsymbol{\theta}} + \left( \frac{\partial \boldsymbol{s}^t}{\partial \boldsymbol{\theta}} + \frac{\partial \boldsymbol{s}^t}{\partial \boldsymbol{y}^{t-1}} \frac{\partial \boldsymbol{y}^{t-1}}{\partial \boldsymbol{\theta}} \right) \right) + \frac{\partial \boldsymbol{y}^t}{\partial \boldsymbol{\theta}} \right]$$

$$= \sum_{1 \leq t \leq T} \mathbf{L}^t \left[ \frac{\partial \boldsymbol{y}^t}{\partial \boldsymbol{s}^t} \boldsymbol{\epsilon}^{t,\boldsymbol{\theta}} + \frac{\partial \boldsymbol{y}^t}{\partial \boldsymbol{\theta}} \right]$$

$$= \sum_{1 \leq t \leq T} \mathbf{L}^t \mathbf{e}^{t,\boldsymbol{\theta}}, \quad (12)$$

where we define the eligibility trace $\mathbf{e}^{t,\boldsymbol{\theta}}$ and the learning signal $\mathbf{L}^t$ as

$$\mathbf{e}^{t,\boldsymbol{\theta}} := \frac{\mathrm{d}\boldsymbol{y}^t}{\mathrm{d}\boldsymbol{\theta}} = \frac{\partial \boldsymbol{y}^t}{\partial \boldsymbol{s}^t} \boldsymbol{\epsilon}^{t,\boldsymbol{\theta}} + \frac{\partial \boldsymbol{y}^t}{\partial \boldsymbol{\theta}} \quad (13)$$

$$\mathbf{L}^t := \frac{\partial E^t}{\partial \boldsymbol{y}^t}. \quad (14)$$

We have shown that $\frac{\mathrm{d}E}{\mathrm{d}\boldsymbol{\theta}}$ can be calculated as the sum of products of a learning signal, Equation 14, and an eligibility trace, Equation 13 without loss of generality. Thus, this proves that for a single layer of recurrently connected SNNs, OSTL provides gradient-equivalence to BPTT. Furthermore, this equivalence is preserved even if this recurrent layer is embedded in a deep network architecture comprising an arbitrary number of stateless layers, e.g., a series of convolutions followed by the recurrent layer and a softmax output layer. Note that the detailed proof of the last statement can be found in Supplementary material S4.

**Derivation of OSTL for deep recurrent networks**

We now turn to the more general case of a multi-layer network, where the state and output equations are as follows

$$s_l^t = \psi(s_l^{t-1}, y_l^{t-1}, y_{l-1}^t, \boldsymbol{\theta}_l) \tag{15}$$

$$y_l^t = \chi(s_l^t, \boldsymbol{\theta}_l), \tag{16}$$

see Section 2 of the main text.

Using this reformulation, Equation 1 is generalized as follows

$$\frac{dE}{d\boldsymbol{\theta}_l} = \sum_{1 \le t \le T} \frac{dE^t}{d\boldsymbol{\theta}_l} = \sum_{1 \le t \le T} \frac{\partial E^t}{\partial y_k^t} \left[ \frac{\partial y_k^t}{\partial s_k^t} \frac{ds_k^t}{d\boldsymbol{\theta}_l} + \frac{\partial y_k^t}{\partial \boldsymbol{\theta}_l} \right]. \tag{17}$$

For the last layer of a multi-layer network, where $l = k$, Equation 17 corresponds to Equation 1 for a single layer. However, for the hidden layers, i.e., $l \ne k$, the term $\frac{ds_k^t}{d\boldsymbol{\theta}_l}$ is expanded as follows

$$\frac{ds_k^t}{d\boldsymbol{\theta}_l} = \frac{\partial s_k^t}{\partial \boldsymbol{\theta}_l} + \frac{\partial s_k^t}{\partial y_k^{t-1}} \frac{dy_k^{t-1}}{d\boldsymbol{\theta}_l} + \frac{\partial s_k^t}{\partial y_{k-1}^t} \frac{dy_{k-1}^t}{d\boldsymbol{\theta}_l}. \tag{18}$$

As one can readily see, there are cross-layer dependencies involved, in particular through the terms $\frac{dy_k^{t-1}}{d\boldsymbol{\theta}_l}$ and $\frac{dy_{k-1}^t}{d\boldsymbol{\theta}_l}$. We define the recursive term $\epsilon_{m,l}^{t,\boldsymbol{\theta}}$ as

$$\begin{aligned}
\epsilon_{m,l}^{t,\boldsymbol{\theta}} &:= \frac{ds_m^t}{d\boldsymbol{\theta}_l} \\
&= \frac{\partial s_m^t}{\partial s_m^{t-1}} \epsilon_{m,l}^{t-1,\boldsymbol{\theta}} + \frac{\partial s_m^t}{\partial y_m^{t-1}} \left( \frac{\partial y_m^{t-1}}{\partial s_m^{t-1}} \epsilon_{m,l}^{t-1,\boldsymbol{\theta}} + \frac{\partial y_m^{t-1}}{\partial \boldsymbol{\theta}_l} \right) \\
&\quad + \frac{\partial s_m^t}{\partial y_{m-1}^t} \left( \frac{\partial y_{m-1}^t}{\partial s_{m-1}^t} \epsilon_{m-1,l}^{t,\boldsymbol{\theta}} + \frac{\partial y_{m-1}^t}{\partial \boldsymbol{\theta}_l} \right) + \frac{\partial s_m^t}{\partial \boldsymbol{\theta}_l},
\end{aligned} \tag{19}$$

with the following properties

$$\begin{aligned}
\epsilon_l^{t,\boldsymbol{\theta}} := \epsilon_{l,l}^{t,\boldsymbol{\theta}} &= \frac{\partial s_l^t}{\partial s_l^{t-1}} \epsilon_l^{t-1,\boldsymbol{\theta}} + \frac{\partial s_l^t}{\partial y_l^{t-1}} \left( \frac{\partial y_l^{t-1}}{\partial s_l^{t-1}} \epsilon_l^{t-1,\boldsymbol{\theta}} + \frac{\partial y_l^{t-1}}{\partial \boldsymbol{\theta}_l} \right) + \frac{\partial s_l^t}{\partial \boldsymbol{\theta}_l} \\
&= \left( \frac{ds_l^t}{ds_l^{t-1}} \epsilon_l^{t-1,\boldsymbol{\theta}} + \left( \frac{\partial s_l^t}{\partial \boldsymbol{\theta}_l} + \frac{\partial s_l^t}{\partial y_l^{t-1}} \frac{\partial y_l^{t-1}}{\partial \boldsymbol{\theta}_l} \right) \right)
\end{aligned} \tag{20}$$

$$e_l^{t,\boldsymbol{\theta}} := \frac{dy_l^t}{d\boldsymbol{\theta}_l} = \left( \frac{\partial y_l^t}{\partial s_l^t} \epsilon_l^{t,\boldsymbol{\theta}} + \frac{\partial y_l^t}{\partial \boldsymbol{\theta}_l} \right) \tag{21}$$

$$\epsilon_{m,l}^{t<1,\boldsymbol{\theta}} = \mathbf{0}$$

$$\epsilon_{l,l+1}^{t,\boldsymbol{\theta}} = \mathbf{0}$$

$$\epsilon_{m,l<1}^{t,\boldsymbol{\theta}} = \mathbf{0}$$

$$\epsilon_{m<1,l}^{t,\boldsymbol{\theta}} = \mathbf{0}.$$

The term $\epsilon_{m,l}^{t,\boldsymbol{\theta}}$ for $m \ne l$ and $l \ne k$ contains a recursion in time, i.e. it depends on $\epsilon_{m,l}^{t-1,\boldsymbol{\theta}}$, similarly as Equation 8, but additionally it contains a recursion in space, i.e. it depends on other layers through $\epsilon_{m-1,l}^{t,\boldsymbol{\theta}}$.

If we insert the term $\epsilon_{m,l}^{t,\boldsymbol{\theta}}$ from Equation 19 into Equation 17

$$\frac{dE}{d\boldsymbol{\theta}_l} = \sum_{1 \le t \le T} \frac{\partial E^t}{\partial y_k^t} \left[ \frac{\partial y_k^t}{\partial s_k^t} \epsilon_{k,l}^{t,\boldsymbol{\theta}} + \frac{\partial y_k^t}{\partial \boldsymbol{\theta}_l} \right], \tag{22}$$

and expand the resulting equation recursively over one step

$$\frac{\mathrm{d}E}{\mathrm{d}\boldsymbol{\theta}_l} = \sum_{1\leq t \leq T} \left[ \frac{\mathrm{d}E^t}{\mathrm{d}\boldsymbol{y}_k^t} \left[ \frac{\partial \boldsymbol{y}_k^t}{\partial \boldsymbol{s}_k^t} \frac{\partial \boldsymbol{s}_k^t}{\partial \boldsymbol{s}_k^{t-1}} \boldsymbol{\epsilon}_{k,l}^{t-1,\boldsymbol{\theta}} + \frac{\partial \boldsymbol{y}_k^t}{\partial \boldsymbol{s}_k^t} \frac{\partial \boldsymbol{s}_k^t}{\partial \boldsymbol{y}_k^{t-1}} \left( \frac{\partial \boldsymbol{y}_k^{t-1}}{\partial \boldsymbol{s}_k^{t-1}} \boldsymbol{\epsilon}_{k,l}^{t-1,\boldsymbol{\theta}} + \frac{\partial \boldsymbol{y}_k^{t-1}}{\partial \boldsymbol{\theta}_l} \right) \right.\right.$$
$$\left. + \frac{\partial \boldsymbol{y}_k^t}{\partial \boldsymbol{s}_k^t} \frac{\partial \boldsymbol{s}_k^t}{\partial \boldsymbol{y}_{k-1}^t} \left( \frac{\partial \boldsymbol{y}_{k-1}^t}{\partial \boldsymbol{s}_{k-1}^t} \boldsymbol{\epsilon}_{k-1,l}^{t,\boldsymbol{\theta}} + \frac{\partial \boldsymbol{y}_{k-1}^t}{\partial \boldsymbol{\theta}_l} \right) + \frac{\partial \boldsymbol{y}_k^t}{\partial \boldsymbol{s}_k^t} \frac{\partial \boldsymbol{s}_k^t}{\partial \boldsymbol{\theta}_l} \right]$$
$$\left. + \frac{\partial \boldsymbol{y}_k^t}{\partial \boldsymbol{\theta}_l} \right], \tag{23}$$

the two recurrencies — $\boldsymbol{\epsilon}_{k-1,l}^{t,\boldsymbol{\theta}}$ in space, and $\boldsymbol{\epsilon}_{k,l}^{t-1,\boldsymbol{\theta}}$ in time — become apparent. Therefore, it can be seen that when expanding $\boldsymbol{\epsilon}_{k-1,l}^t$ far enough in space, eventually terms involving $\boldsymbol{\epsilon}_{l,l}^{t,\boldsymbol{\theta}} = \boldsymbol{\epsilon}_l^t$ are reached. We isolate the components involving only layer $l$, which allows us to rewrite Equation 23 as

$$\frac{\mathrm{d}E}{\mathrm{d}\boldsymbol{\theta}_l} = \sum_{1 \leq t \leq T} \left[ \frac{\mathrm{d}E^t}{\mathrm{d}\boldsymbol{y}_k^t} \left( \prod_{(k-l+1)>m\geq 1} \frac{\partial \boldsymbol{y}_{k-m+1}^t}{\partial \boldsymbol{s}_{k-m+1}^t} \frac{\partial \boldsymbol{s}_{k-m+1}^t}{\partial \boldsymbol{y}_{k-m}^t} \right) \left( \frac{\partial \boldsymbol{y}_l^t}{\partial \boldsymbol{s}_l^t} \boldsymbol{\epsilon}_l^{t,\boldsymbol{\theta}} + \frac{\partial \boldsymbol{y}_l^t}{\partial \boldsymbol{\theta}_l} \right) + \mathbf{R} \right]. \tag{24}$$

Note that all remaining terms were collected into a residual term $\mathbf{R}$. In addition, we define a generalized learning signal $\mathbf{L}_l^t$ and a generalized eligibility trace $\mathbf{e}_l^{t,\boldsymbol{\theta}}$ as

$$\mathbf{L}_l^t = \frac{\partial E^t}{\partial \boldsymbol{y}_k^t} \left( \prod_{(k-l+1)\geq m\geq 1} \frac{\partial \boldsymbol{y}_{k-m+1}^t}{\partial \boldsymbol{s}_{k-m+1}^t} \frac{\partial \boldsymbol{s}_{k-m+1}^t}{\partial \boldsymbol{y}_{k-m}^t} \right) \tag{25}$$

$$\mathbf{e}_l^{t,\boldsymbol{\theta}} = \left( \frac{\partial \boldsymbol{y}_l^t}{\partial \boldsymbol{s}_l^t} \boldsymbol{\epsilon}_l^{t,\boldsymbol{\theta}} + \frac{\partial \boldsymbol{y}_l^t}{\partial \boldsymbol{\theta}_l} \right), \tag{26}$$

see Equations 8 and 9 of the main text. This allows to express the parameter update as

$$\frac{\mathrm{d}E}{\mathrm{d}\boldsymbol{\theta}_l} = \sum_{1 \leq t \leq T} \left[ \mathbf{L}_l^t \mathbf{e}_l^{t,\boldsymbol{\theta}} + \mathbf{R} \right], \tag{27}$$

see Equation 7 of the main text.

Finally, by omitting the residual term $\mathbf{R}$, we arrive at the simplified approximate expression of OSTL for deep recurrent networks, given in Equation 11 of the main text.

## Supplementary material S2: Complexity analysis

In this section we investigate the computational complexity of OSTL for a single SNN layer, where the state and output equations are given by

$$\boldsymbol{s}_l^t = \mathbf{g}(\mathbf{W}_l \boldsymbol{y}_{l-1}^t + \mathbf{H}_l \boldsymbol{y}_l^{t-1} + d \cdot \boldsymbol{s}_l^{t-1} \odot (\mathbb{1} - \boldsymbol{y}_l^{t-1})) \tag{28}$$
$$\boldsymbol{y}_l^t = \mathbf{h}(\boldsymbol{s}_l^t + \mathbf{b}_l) \tag{29}$$

see Equations 1 and 2 of the main text. The majority of the computations arise from calculating the eligibility vectors as given in Equation 10. Specifically, the term $\frac{\mathrm{d}\boldsymbol{s}^t}{\mathrm{d}\boldsymbol{s}^{t-1}}\boldsymbol{\epsilon}^{t-1,\boldsymbol{\theta}}$ is computationally the most intensive one, as it requires matrix multiplications. We use the index notation to explicitly investigate the complexity in terms of the number of multiplications involved. Note that the indices do not indicate a specific layer, but indicate elements of a matrix. The eligibility vectors are computed as

$$\epsilon_{opq}^{t,\boldsymbol{\theta}} = \left( \frac{\mathrm{d}s_o^t}{\mathrm{d}s_r^{t-1}} \epsilon_{rpq}^{t-1,\boldsymbol{\theta}} + \left( \frac{\partial s_o^t}{\partial \theta_{pq}} + \frac{\partial s_o^t}{\partial y_r^{t-1}} \frac{\partial y_r^{t-1}}{\partial \theta_{pq}} \right) \right) \tag{30}$$
$$\epsilon_{opq}^{t,\boldsymbol{\theta}} \sim \frac{\mathrm{d}s_o^t}{\mathrm{d}s_r^{t-1}} \epsilon_{rpq}^{t-1,\boldsymbol{\theta}}.$$

Thus, we observe

$$\epsilon_{opq}^{t,\boldsymbol{\theta}} \sim O(n)$$
$$\boldsymbol{\epsilon}^{t,\boldsymbol{\theta}} \sim O(n^4). \tag{31}$$

In particular, computing a single value of the eligibility vector $\epsilon_{opq}^{t,\boldsymbol{\theta}}$ requires $O(n)$ multiplications, where $n$ denotes the number of units in the layer. Therefore, the total computational cost of OSTL for shallow SNNs is $O(n^4)$ as stated in Table 1 of the main text. In case of a multi-layered architecture, this complexity scales with the number of layers, i.e., $O(kn^4)$, where $k$ denotes the number of layers and $n = \max_{1 \leq l \leq k} n_l$ denotes the maximum layer size in the network.

However, if we analyze a feed-forward SNN with the state and output equations

$$\boldsymbol{s}_l^t = \mathbf{g}(\mathbf{W}_l \boldsymbol{y}_{l-1}^t + d \cdot \boldsymbol{s}_l^{t-1} \odot (\mathbb{1} - \boldsymbol{y}_l^{t-1})) \tag{32}$$
$$\boldsymbol{y}_l^t = \mathbf{h}(\boldsymbol{s}_l^t + \mathbf{b}_l), \tag{33}$$

the eligibility vector $\epsilon_{opq}^{t,\boldsymbol{\theta}}$ reduces to a rank-two tensor, because the Jacobian $\frac{\mathrm{d}\boldsymbol{s}_o^t}{\mathrm{d}\boldsymbol{s}_r^{t-1}}$ reduces to a diagonal matrix as

$$\epsilon_{opq}^{t,\boldsymbol{\theta}} = \left( \frac{\mathrm{d}s_o^t}{\mathrm{d}s_r^{t-1}} \delta_{or} \epsilon_{rpq}^{t-1,\boldsymbol{\theta}} + \left( \frac{\partial s_o^t}{\partial \theta_{pq}} \delta_{oq} + \frac{\partial s_o^t}{\partial y_r^{t-1}} \delta_{or} \frac{\partial y_r^{t-1}}{\partial \theta_{pq}} \delta_{rq} \right) \right) \tag{34}$$
$$\epsilon_{op}^{t,\boldsymbol{\theta}} \sim \frac{\mathrm{d}s_o^t}{\mathrm{d}s_r^{t-1}} \delta_{or} \epsilon_{rp}^{t-1,\boldsymbol{\theta}}$$
$$\sim \frac{\mathrm{d}s_o^t}{\mathrm{d}s_o^{t-1}} \epsilon_{op}^{t-1,\boldsymbol{\theta}}. \tag{35}$$

Thus, we observe that the number of multiplications is

$$\epsilon_{op}^{t,\boldsymbol{\theta}} \sim O(1) \tag{36}$$
$$\boldsymbol{\epsilon}^{t,\boldsymbol{\theta}} \sim O(n^2). \tag{37}$$

This implies that OSTL enables training of feed-forward SNNs with reduced time complexity of $O(n^2)$ for shallow and of $O(kn^2)$ for $k$-layered networks. Note that the aforementioned analysis holds also for single or multiple layers of generic RNNs.

## Supplementary material S3: Extensions of OSTL with biological constraints

To derive OSTL, we revisited the formulation of gradient-based training and incorporated insights from biological systems. In particular, we separated the gradient components into spatial and temporal gradients, corresponding to biologically-inspired notions of learning signals and eligibility traces. As already indicated in the main text, in OSTL the eligibility trace involves only information local to a layer of neurons, while the learning signal relies on the mathematical concept of error backpropagation applied locally in time.

OSTL may be considered of as the baseline algorithm with exact derivations for the temporal and spatial components. It is possible to further study and introduce approximations into both components, for example to evaluate the impact of introducing more stringent biological constraints. In the following, we discuss a few examples of such approaches, which also highlight the benefits of the clear separation of the gradient components. Note that such approximations can be independently introduced to the temporal and/or the spatial gradients.

For example, approximate eligibility traces may be used. Their exact formulation for recurrent network architectures involves computations with the recurrent matrix $\mathbf{H}_l$, see Equation 20-23 of the main text, which requires information that is local to one layer, but non-local to an individual neuron. In addition, these computations dominate to a large extent the time complexity of OSTL, as discussed in detail in Supplementary material S2. To remove the need to transmit this information, one could neglect the influence of the recurrent matrix $\mathbf{H}$ in the eligibility traces, leading to an approximation of OSTL (OSTL w/o $\mathbf{H}$), similar to [1]. Although this scheme may lead to a performance degradation in certain tasks, it substantially reduces the time complexity of OSTL for generic RNNs from $O(kn^4)$ down to $O(kn^2)$, see Supplementary material S2.

Further biological constraints may be introduced into the spatial gradient components. Specifically, the learning signal $\mathbf{L}_l^t$ of OSTL, as defined in Equation 9 of the main text, uses the same weight matrix to transport information in the forward pass for time step $t$ as well as for the spatial credit assignment at this time step – $W_{l+1}$ in Equation 18 of the main text. A common concern in computational neuroscience is that signals in biological substrate should not use the same synaptic connection in both directions. To address this concern, OSTL can utilize schemes such as feedback alignment [2], direct feedback alignment or indirect feedback alignment [3]. In feedback alignment, the learning signals are delivered via separate connections and random matrices are used to deliver the learning signal from one layer to the previous one. Considering the mean squared error loss with readout weights given in Equation 17 of the main text

$$E^t = \frac{1}{2} \sum_{n_k} \left( \mathbf{W}^{\text{out}} \boldsymbol{y}_k^t - \hat{\boldsymbol{y}}^t \right)^2, \tag{38}$$

the learning signals become

$$\mathbf{L}_l^t = \mathbf{L}_{l+1}^t \left( \text{diag} \left( \mathbf{h}'^t_{l+1} \odot \mathbf{g}'^t_{l+1} \right) \mathbf{B}_{l+1} \right), \tag{39}$$

with

$$\mathbf{L}_k^t = \left( \left( \mathbf{W}^{\text{out}} \right)^T \boldsymbol{y}_k^t - \hat{\boldsymbol{y}}^t \right), \tag{40}$$

where $\mathbf{B}_{l+1}$ is a random matrix connecting the layer $l+1$ to $l$. Because of the random nature of this approximation, we refer to this algorithm as OSTL rnd.

To demonstrate the performance of the introduced approximations, we evaluated them along with regular OSTL on the speech recognition task TIMIT and the performance is summarized in Supplementary Table 1. The task and network architecture are described in Section 4 of the main text with the details given in Supplementary material S7.

The aforementioned examples highlight that OSTL allows to easily incorporate more biological constraints and thereby even bridge different approaches from literature.

Table 1: Extended performance comparison for the TIMIT dataset.

| Unit | Architecture | Algorithm | Error rate |
|---|---|---|---|
| sSNU | 1×200 | OSTL | 24.58% ± 0.03 |
| sSNU | 1×200 | OSTL w/o **H** | 28.45% ± 0.12 |
| sSNU | 1×200 | OSTL rnd | 31.82% ± 0.20 |
| sSNU | 1×200 | OSTL rnd + w/o **H** | 41.38% ± 3.40 |

## Supplementary material S4: OSTL applied to recurrent layers embedded in stateless networks

In many network architectures, a recurrent layer is embedded along with an arbitrary number of non-recurrent, stateless layers, for example a softmax output layer. From the perspective of OSTL, a stateless layer does not introduce any residual terms $R$ and therefore OSTL maintains gradient equivalence with BPTT even for deep architectures. To demonstrate this, we consider a deep network consisting of stateless layers, with the following state and output equations

$$\boldsymbol{s}_l^t = \boldsymbol{\psi}(\boldsymbol{y}_{l-1}^t, \boldsymbol{\theta}_l) \tag{41}$$

$$\boldsymbol{y}_l^t = \boldsymbol{\chi}(\boldsymbol{s}_l^t, \boldsymbol{\theta}_l). \tag{42}$$

These equations simplify the term $\boldsymbol{\epsilon}_{m,l}^{t,\boldsymbol{\theta}}$ from Equation 19 to:

$$\boldsymbol{\epsilon}_{m,l}^{t,\boldsymbol{\theta}} = \frac{\partial \boldsymbol{s}_m^t}{\partial \boldsymbol{y}_{m-1}^t} \left( \frac{\partial \boldsymbol{y}_{m-1}^t}{\partial \boldsymbol{s}_{m-1}^t} \boldsymbol{\epsilon}_{m-1,l}^{t,\boldsymbol{\theta}} + \frac{\partial \boldsymbol{y}_{m-1}^t}{\partial \boldsymbol{\theta}_l} \right) + \frac{\partial \boldsymbol{s}_m^t}{\partial \boldsymbol{\theta}_l}. \tag{43}$$

If we insert this term to Equation 22, we obtain

$$\frac{\mathrm{d}E}{\mathrm{d}\boldsymbol{\theta}_l} = \sum_{1 \leq t \leq T} \left[ \frac{\mathrm{d}E^t}{\mathrm{d}\boldsymbol{y}_k^t} \left[ \frac{\partial \boldsymbol{y}_k^t}{\partial \boldsymbol{s}_k^t} \frac{\partial \boldsymbol{s}_k^t}{\partial \boldsymbol{y}_{k-1}^t} \left( \frac{\partial \boldsymbol{y}_{k-1}^t}{\partial \boldsymbol{s}_{k-1}^t} \boldsymbol{\epsilon}_{k-1}^{t,\boldsymbol{\theta}} + \frac{\partial \boldsymbol{y}_{k-1}^t}{\partial \boldsymbol{\theta}_l} \right) + \frac{\partial \boldsymbol{y}_k^t}{\partial \boldsymbol{s}_k^t} \frac{\partial \boldsymbol{s}_k^t}{\partial \boldsymbol{\theta}_l} \right] + \frac{\partial \boldsymbol{y}_k^t}{\partial \boldsymbol{\theta}_l} \right]. \tag{44}$$

By recursively inserting Equation 43 into Equation 44 until $k - r = l$, most terms include partial derivatives of the form $\frac{\partial \boldsymbol{y}_{k-r}^t}{\partial \boldsymbol{\theta}_l}$ or $\frac{\partial \boldsymbol{s}_{k-r}^t}{\partial \boldsymbol{\theta}_l}$. These terms vanish because the layer $k - r$ does not depend on parameters $\boldsymbol{\theta}_l$, except when $k - r = l$. Therefore, Equation 44 can be written as

$$\frac{\mathrm{d}E}{\mathrm{d}\boldsymbol{\theta}_l} = \sum_{1 \leq t \leq T} \left[ \frac{\mathrm{d}E^t}{\mathrm{d}\boldsymbol{y}_k^t} \left( \prod_{(k-l+1) > m \geq 1} \frac{\partial \boldsymbol{y}_{k-m+1}^t}{\partial \boldsymbol{s}_{k-m+1}^t} \frac{\partial \boldsymbol{s}_{k-m+1}^t}{\partial \boldsymbol{y}_{k-m}^t} \right) \left( \frac{\partial \boldsymbol{y}_l^t}{\partial \boldsymbol{s}_l^t} \boldsymbol{\epsilon}_l^{t,\boldsymbol{\theta}} + \frac{\partial \boldsymbol{y}_l^t}{\partial \boldsymbol{\theta}_l} \right) \right], \tag{45}$$

which is the combination of the generalized learning signal $L_l^t$ and eligibility trace $e_l^{t,\boldsymbol{\theta}}$ without any residual term **R**, as stated in Equations 27

$$\frac{\mathrm{d}E}{\mathrm{d}\boldsymbol{\theta}_l} = \sum_{1 \leq t \leq T} \mathbf{L}_l^t \mathbf{e}_l^{t,\boldsymbol{\theta}}. \tag{46}$$

## Supplementary material S5: OSTL for Long Short-Term Memory

Because LSTM units are probably the most commonly used type of RNN units in contemporary state-of-the-art machine learning applications, we include an explicit derivation of OSTL for them. An LSTM unit contains three gates and an internal state. A single layer $l$, composed of LSTM units, is governed by the following equations

$$\mathbf{i}_l^t = \mathbf{g}(\mathbf{W}_l^i \mathbf{y}_{l-1}^t + \mathbf{H}_l^i \mathbf{y}_l^{t-1} + \mathbf{b}_l^i) \tag{47}$$

$$\mathbf{c}_l^t = \mathbf{g}(\mathbf{W}_l^c \mathbf{y}_{l-1}^t + \mathbf{H}_l^c \mathbf{y}_l^{t-1} + \mathbf{b}_l^c) \tag{48}$$

$$\mathbf{f}_l^t = \mathbf{g}(\mathbf{W}_l^f \mathbf{y}_{l-1}^t + \mathbf{H}_l^f \mathbf{y}_l^{t-1} + \mathbf{b}_l^f) \tag{49}$$

$$\mathbf{s}_l^t = \mathbf{f}_l^t \odot \mathbf{s}_l^{t-1} + \mathbf{i}_l^t \odot \mathbf{h}(\mathbf{W}_l^s \mathbf{y}_{l-1}^t + \mathbf{H}_l^s \mathbf{y}_l^{t-1} + \mathbf{b}_l^s) \tag{50}$$

$$= \mathbf{g}(\mathbf{W}_l^f \mathbf{y}_{l-1}^t + \mathbf{H}_l^f \mathbf{y}_l^{t-1} + \mathbf{b}_l^f) \odot \mathbf{s}_l^{t-1} +$$
$$\mathbf{g}(\mathbf{W}_l^i \mathbf{y}_{l-1}^t + \mathbf{H}_l^i \mathbf{y}_l^{t-1} + \mathbf{b}_l^i) \odot \mathbf{h}(\mathbf{W}_l^s \mathbf{y}_{l-1}^t + \mathbf{H}_l^s \mathbf{y}_l^{t-1} + \mathbf{b}_l^s) \tag{51}$$

$$\mathbf{y}_l^t = \mathbf{c}_l^t \odot \mathbf{h}(\mathbf{s}_l^t) \tag{52}$$

$$= \mathbf{g}(\mathbf{W}_l^c \mathbf{y}_{l-1}^t + \mathbf{H}_l^c \mathbf{y}_l^{t-1} + \mathbf{b}_l^c) \odot \mathbf{h}(\mathbf{s}_l^t), \tag{53}$$

where typically $\mathbf{g} = \boldsymbol{\sigma}$ and $\mathbf{h} = \tanh$. Note that we reformulated Equations 51 and 53 such that one can immediately see the resemblance to the generalized Equations 25-26 of the main text.

By using Equation 26, the eligibility traces for a layer of LSTM units take the following form

$$\mathbf{e}_l^{t,\mathbf{W}^i} = \text{diag}\left(\mathbf{c}_l^t \odot \mathbf{h}_l'^{y,t}\right) \boldsymbol{\epsilon}^{t,\mathbf{W}^i} + \text{diag}\left(\mathbf{h}_l^{y,t} \odot \mathbf{g}_l'^{c,t}\right) \mathbf{H}_l^c \mathbf{e}_l^{t-1,\mathbf{W}^i} \tag{54}$$

$$\mathbf{e}_l^{t,\mathbf{H}^i} = \text{diag}\left(\mathbf{c}_l^t \odot \mathbf{h}_l'^{y,t}\right) \boldsymbol{\epsilon}^{t,\mathbf{H}^i} + \text{diag}\left(\mathbf{h}_l^{y,t} \odot \mathbf{g}_l'^{c,t}\right) \mathbf{H}_l^c \mathbf{e}_l^{t-1,\mathbf{H}^i} \tag{55}$$

$$\mathbf{e}_l^{t,\mathbf{b}^i} = \text{diag}\left(\mathbf{c}_l^t \odot \mathbf{h}_l'^{y,t}\right) \boldsymbol{\epsilon}^{t,\mathbf{b}^i} + \text{diag}\left(\mathbf{h}_l^{y,t} \odot \mathbf{g}_l'^{c,t}\right) \mathbf{H}_l^c \mathbf{e}_l^{t-1,\mathbf{b}^i} \tag{56}$$

$$\mathbf{e}_l^{t,\mathbf{W}^f} = \text{diag}\left(\mathbf{c}_l^t \odot \mathbf{h}_l'^{y,t}\right) \boldsymbol{\epsilon}^{t,\mathbf{W}^f} + \text{diag}\left(\mathbf{h}_l^{y,t} \odot \mathbf{g}_l'^{c,t}\right) \mathbf{H}_l^c \mathbf{e}_l^{t-1,\mathbf{W}^f} \tag{57}$$

$$\mathbf{e}_l^{t,\mathbf{H}^f} = \text{diag}\left(\mathbf{c}_l^t \odot \mathbf{h}_l'^{y,t}\right) \boldsymbol{\epsilon}^{t,\mathbf{H}^f} + \text{diag}\left(\mathbf{h}_l^{y,t} \odot \mathbf{g}_l'^{c,t}\right) \mathbf{H}_l^c \mathbf{e}_l^{t-1,\mathbf{H}^f} \tag{58}$$

$$\mathbf{e}_l^{t,\mathbf{b}^f} = \text{diag}\left(\mathbf{c}_l^t \odot \mathbf{h}_l'^{y,t}\right) \boldsymbol{\epsilon}^{t,\mathbf{b}^f} + \text{diag}\left(\mathbf{h}_l^{y,t} \odot \mathbf{g}_l'^{c,t}\right) \mathbf{H}_l^c \mathbf{e}_l^{t-1,\mathbf{b}^f} \tag{59}$$

$$\mathbf{e}_l^{t,\mathbf{W}^s} = \text{diag}\left(\mathbf{c}_l^t \odot \mathbf{h}_l'^{y,t}\right) \boldsymbol{\epsilon}^{t,\mathbf{W}^s} + \text{diag}\left(\mathbf{h}_l^{y,t} \odot \mathbf{g}_l'^{c,t}\right) \mathbf{H}_l^c \mathbf{e}_l^{t-1,\mathbf{W}^s} \tag{60}$$

$$\mathbf{e}_l^{t,\mathbf{H}^s} = \text{diag}\left(\mathbf{c}_l^t \odot \mathbf{h}_l'^{y,t}\right) \boldsymbol{\epsilon}^{t,\mathbf{H}^s} + \text{diag}\left(\mathbf{h}_l^{y,t} \odot \mathbf{g}_l'^{c,t}\right) \mathbf{H}_l^c \mathbf{e}_l^{t-1,\mathbf{H}^s} \tag{61}$$

$$\mathbf{e}_l^{t,\mathbf{b}^s} = \text{diag}\left(\mathbf{c}_l^t \odot \mathbf{h}_l'^{y,t}\right) \boldsymbol{\epsilon}^{t,\mathbf{b}^s} + \text{diag}\left(\mathbf{h}_l^{y,t} \odot \mathbf{g}_l'^{c,t}\right) \mathbf{H}_l^c \mathbf{e}_l^{t-1,\mathbf{b}^s} \tag{62}$$

$$\mathbf{e}_l^{t,\mathbf{W}^c} = \text{diag}\left(\mathbf{c}_l^t \odot \mathbf{h}_l'^{y,t}\right) \boldsymbol{\epsilon}^{t,\mathbf{W}^c} + \text{diag}\left(\mathbf{h}_l^{y,t} \odot \mathbf{g}_l'^{c,t}\right) \boldsymbol{\Upsilon}_{l-1}^t + \text{diag}\left(\mathbf{h}_l^{y,t} \odot \mathbf{g}_l'^{c,t}\right) \mathbf{H}_l^c \mathbf{e}_l^{t-1,\mathbf{W}^c} \tag{63}$$

$$\mathbf{e}_l^{t,\mathbf{H}^c} = \text{diag}\left(\mathbf{c}_l^t \odot \mathbf{h}_l'^{y,t}\right) \boldsymbol{\epsilon}^{t,\mathbf{H}^c} + \text{diag}\left(\mathbf{h}_l^{y,t} \odot \mathbf{g}_l'^{c,t}\right) \boldsymbol{\Upsilon}_l^{t-1} + \text{diag}\left(\mathbf{h}_l^{y,t} \odot \mathbf{g}_l'^{c,t}\right) \mathbf{H}_l^c \mathbf{e}_l^{t-1,\mathbf{H}^c} \tag{64}$$

$$\mathbf{e}_l^{t,\mathbf{b}^c} = \text{diag}\left(\mathbf{c}_l^t \odot \mathbf{h}_l'^{y,t}\right) \boldsymbol{\epsilon}^{t,\mathbf{b}^c} + \text{diag}\left(\mathbf{h}_l^{y,t} \odot \mathbf{g}_l'^{c,t}\right) + \text{diag}\left(\mathbf{h}_l^{y,t} \odot \mathbf{g}_l'^{c,t}\right) \mathbf{H}_l^c \mathbf{e}_l^{t-1,\mathbf{b}^c}, \tag{65}$$

using Equation 10 the eligibility vectors in Equations 54-65 become

$$\epsilon_l^{t,\mathbf{W}^i} = \frac{\mathrm{d}\boldsymbol{s}_l^t}{\mathrm{d}\boldsymbol{s}_l^{t-1}}\epsilon_l^{t-1,\mathbf{W}^i} + \mathrm{diag}\left(\mathbf{h}_l^{s,t}\odot\mathbf{g}'^{i,t}_l\right)\boldsymbol{\Upsilon}_{l-1}^t \tag{66}$$

$$\epsilon_l^{t,\mathbf{H}^i} = \frac{\mathrm{d}\boldsymbol{s}_l^t}{\mathrm{d}\boldsymbol{s}_l^{t-1}}\epsilon_l^{t-1,\mathbf{H}^i} + \mathrm{diag}\left(\mathbf{h}_l^{s,t}\odot\mathbf{g}'^{i,t}_l\right)\boldsymbol{\Upsilon}_l^{t-1} \tag{67}$$

$$\epsilon_l^{t,\mathbf{b}^i} = \frac{\mathrm{d}\boldsymbol{s}_l^t}{\mathrm{d}\boldsymbol{s}_l^{t-1}}\epsilon_l^{t-1,\mathbf{b}^i} + \mathrm{diag}\left(\mathbf{h}_l^{s,t}\odot\mathbf{g}'^{i,t}_l\right) \tag{68}$$

$$\epsilon_l^{t,\mathbf{W}^f} = \frac{\mathrm{d}\boldsymbol{s}_l^t}{\mathrm{d}\boldsymbol{s}_l^{t-1}}\epsilon_l^{t-1,\mathbf{W}^f} + \mathrm{diag}\left(\boldsymbol{s}_l^{t-1}\odot\mathbf{g}'^{f,t}_l\right)\boldsymbol{\Upsilon}_{l-1}^t \tag{69}$$

$$\epsilon_l^{t,\mathbf{H}^f} = \frac{\mathrm{d}\boldsymbol{s}_l^t}{\mathrm{d}\boldsymbol{s}_l^{t-1}}\epsilon_l^{t-1,\mathbf{H}^f} + \mathrm{diag}\left(\boldsymbol{s}_l^{t-1}\odot\mathbf{g}'^{f,t}_l\right)\boldsymbol{\Upsilon}_l^{t-1} \tag{70}$$

$$\epsilon_l^{t,\mathbf{b}^f} = \frac{\mathrm{d}\boldsymbol{s}_l^t}{\mathrm{d}\boldsymbol{s}_l^{t-1}}\epsilon_l^{t-1,\mathbf{b}^f} + \mathrm{diag}\left(\boldsymbol{s}_l^{t-1}\odot\mathbf{g}'^{f,t}_l\right) \tag{71}$$

$$\epsilon_l^{t,\mathbf{W}^s} = \frac{\mathrm{d}\boldsymbol{s}_l^t}{\mathrm{d}\boldsymbol{s}_l^{t-1}}\epsilon_l^{t-1,\mathbf{W}^s} + \mathrm{diag}\left(\mathbf{i}_l^t\odot\mathbf{h}'^{s,t}_l\right)\boldsymbol{\Upsilon}_{l-1}^t \tag{72}$$

$$\epsilon_l^{t,\mathbf{H}^s} = \frac{\mathrm{d}\boldsymbol{s}_l^t}{\mathrm{d}\boldsymbol{s}_l^{t-1}}\epsilon_l^{t-1,\mathbf{H}^s} + \mathrm{diag}\left(\mathbf{i}_l^t\odot\mathbf{h}'^{s,t}_l\right)\boldsymbol{\Upsilon}_l^{t-1} \tag{73}$$

$$\epsilon_l^{t,\mathbf{b}^s} = \frac{\mathrm{d}\boldsymbol{s}_l^t}{\mathrm{d}\boldsymbol{s}_l^{t-1}}\epsilon_l^{t-1,\mathbf{b}^s} + \mathrm{diag}\left(\mathbf{i}_l^t\odot\mathbf{h}'^{s,t}_l\right) \tag{74}$$

$$\begin{aligned}\epsilon_l^{t,\mathbf{W}^c} = &\frac{\mathrm{d}\boldsymbol{s}_l^t}{\mathrm{d}\boldsymbol{s}_l^{t-1}}\epsilon_l^{t-1,\mathbf{W}^c} + \\ &\left(\mathrm{diag}\left(\mathbf{g}'^{f,t}_l\odot\boldsymbol{s}_l^{t-1}\right)\mathbf{H}_l^f\mathrm{diag}\left(\mathbf{g}'^{c,t-1}_l\odot\mathbf{h}_l^{y,t-1}\right)\boldsymbol{\Upsilon}_{l-1}^{t-1}\right.\\ &+\mathrm{diag}\left(\mathbf{h}_l^{s,t}\odot\mathbf{g}'^{i,t}_l\right)\mathbf{H}_l^i\mathrm{diag}\left(\mathbf{g}'^{c,t-1}_l\odot\mathbf{h}_l^{y,t-1}\right)\boldsymbol{\Upsilon}_{l-1}^{t-1}\\ &\left.+\mathrm{diag}\left(\mathbf{i}_l^t\odot\mathbf{h}'^{s,t}_l\right)\mathbf{H}_l^s\mathrm{diag}\left(\mathbf{g}'^{c,t-1}_l\odot\mathbf{h}_l^{y,t-1}\right)\boldsymbol{\Upsilon}_{l-1}^{t-1}\right) \end{aligned} \tag{75}$$

$$\begin{aligned}\epsilon_l^{t,\mathbf{H}^c} = &\frac{\mathrm{d}\boldsymbol{s}_l^t}{\mathrm{d}\boldsymbol{s}_l^{t-1}}\epsilon_l^{t-1,\mathbf{H}^c} + \\ &\left(\mathrm{diag}\left(\mathbf{g}'^{f,t}_l\odot\boldsymbol{s}_l^{t-1}\right)\mathbf{H}_l^f\mathrm{diag}\left(\mathbf{g}'^{c,t-1}_l\odot\mathbf{h}_l^{y,t-1}\right)\boldsymbol{\Upsilon}_l^{t-2}\right.\\ &+\mathrm{diag}\left(\mathbf{h}_l^{s,t}\odot\mathbf{g}'^{i,t}_l\right)\mathbf{H}_l^i\mathrm{diag}\left(\mathbf{g}'^{c,t-1}_l\odot\mathbf{h}_l^{y,t-1}\right)\boldsymbol{\Upsilon}_l^{t-2}\\ &\left.+\mathrm{diag}\left(\mathbf{i}_l^t\odot\mathbf{h}'^{s,t}_l\right)\mathbf{H}_l^s\mathrm{diag}\left(\mathbf{g}'^{c,t-1}_l\odot\mathbf{h}_l^{y,t-1}\right)\boldsymbol{\Upsilon}_l^{t-2}\right) \end{aligned} \tag{76}$$

$$\begin{aligned}\epsilon_l^{t,\mathbf{b}^c} = &\frac{\mathrm{d}\boldsymbol{s}_l^t}{\mathrm{d}\boldsymbol{s}_l^{t-1}}\epsilon_l^{t-1,\mathbf{b}^c} + \\ &\left(\mathrm{diag}\left(\mathbf{g}'^{f,t}_l\odot\boldsymbol{s}_l^{t-1}\right)\mathbf{H}_l^f\mathrm{diag}\left(\mathbf{g}'^{c,t-1}_l\odot\mathbf{h}_l^{y,t-1}\right)\right.\\ &+\mathrm{diag}\left(\mathbf{h}_l^{s,t}\odot\mathbf{g}'^{i,t}_l\right)\mathbf{H}_l^i\mathrm{diag}\left(\mathbf{g}'^{c,t-1}_l\odot\mathbf{h}_l^{y,t-1}\right)\\ &\left.+\mathrm{diag}\left(\mathbf{i}_l^t\odot\mathbf{h}'^{s,t}_l\right)\mathbf{H}_l^s\mathrm{diag}\left(\mathbf{g}'^{c,t-1}_l\odot\mathbf{h}_l^{y,t-1}\right)\right), \end{aligned} \tag{77}$$

with

$$\frac{\mathrm{d}\boldsymbol{s}_l^t}{\mathrm{d}\boldsymbol{s}_l^{t-1}} = \mathrm{diag}(\mathbf{f}^t) + \mathrm{diag}\left(\boldsymbol{s}_l^{t-1}\odot\mathbf{g}'^{f,t}_l\right)\mathbf{H}_l^f\mathrm{diag}\left(\mathbf{c}_l^{t-1}\odot\mathbf{h}'^{y,t-1}_l\right) \tag{78}$$

$$+ \mathrm{diag}\left(\mathbf{h}_l^{s,t}\odot\mathbf{g}'^{i,t}_l\right)\mathbf{H}_l^i\mathrm{diag}\left(\mathbf{c}_l^{t-1}\odot\mathbf{h}'^{y,t-1}_l\right) \tag{79}$$

$$+ \mathrm{diag}\left(\mathbf{i}_l^t\odot\mathbf{h}'^{s,t}_l\right)\mathbf{H}_l^s\mathrm{diag}\left(\mathbf{c}_l^{t-1}\odot\mathbf{h}'^{y,t-1}_l\right). \tag{80}$$

We have used the short-hand notation of $\frac{d\psi_l^t}{d\omega^t} = \mathbf{g}'|_{\psi_l^t} = \mathbf{g}'^{\psi,t}_l$ with $\psi = \{\mathbf{i}, \mathbf{c}, \mathbf{f}, \mathbf{s}\}$ and $\omega = \{\mathbf{y}_l^{t-1}, \boldsymbol{\theta}_l\}$, $h(\mathbf{s}_l^t) = \mathbf{h}_l^{y,t}$, $\frac{dh(\mathbf{s}_l^t)}{d\mathbf{s}_l^t} = \mathbf{h}'^{y,t}_l$, $h(\mathbf{W}_l^s \mathbf{y}_{l-1}^t + \mathbf{H}_l^s \mathbf{y}_l^{t-1} + \mathbf{b}_l^s) = \mathbf{h}_l^{s,t}$ and $\frac{d h(\mathbf{W}_l^s \mathbf{y}_{l-1}^t + \mathbf{H}_l^s \mathbf{y}_l^{t-1} + \mathbf{b}_l^s)}{d\omega} = \mathbf{h}'|_{\mathbf{W}_l^s \mathbf{y}_{l-1}^t + \mathbf{H}_l^s \mathbf{y}_l^{t-1} + \mathbf{b}_l^s} = \mathbf{h}'^{y,t}_l$. Furthermore, the elements of $\boldsymbol{\Upsilon}_l^t$ are given by $(\boldsymbol{\Upsilon}_l^t)_{opq} = \delta_{op} (y_l^t)_q$.

Considering the mean squared error loss with readout weights given in Equation 17 of the main text

$$E^t = \frac{1}{2} \sum_{n_k} \left( \mathbf{W}^{\text{out}} \mathbf{y}_k^t - \hat{\mathbf{y}}^t \right)^2, \tag{81}$$

the learning signals become

$$\mathbf{L}_l^t = \mathbf{L}_{l+1}^t \left( \text{diag} \left( \mathbf{c}_{l+1}^t \odot \mathbf{h}'^{y,t}_{l+1} \right) \left[ \text{diag} \left( \mathbf{s}_{l+1}^{t-1} \odot \mathbf{g}'^{f,t}_{l+1} \right) \mathbf{W}_{l+1}^f \right. \right.$$
$$+ \text{diag} \left( \mathbf{h}_{l+1}^{s,t} \odot \mathbf{g}'^{i,t}_{l+1} \right) \mathbf{W}_{l+1}^i + \text{diag} \left( \mathbf{g}_{l+1}^{i,t} \odot \mathbf{h}'^{s,t}_{l+1} \right) \mathbf{W}_{l+1}^s \right]$$
$$\left. + \text{diag} \left( \mathbf{h}_{l+1}^{s,t} \odot \mathbf{g}'^{c,t}_{l+1} \right) \mathbf{H}_{l+1}^c \right), \tag{82}$$

with

$$\mathbf{L}_k^t = \left( \left( \mathbf{W}^{\text{out}} \right)^T \mathbf{y}_k^t - \hat{\mathbf{y}}^t \right). \tag{83}$$

## Supplementary material S6: OSTL for Gated Recurrent Unit

OSTL can also be applied to GRUs. To illustrate this, we start from the state equations

$$\mathbf{u}_l^t = \mathbf{g}(\mathbf{W}_l^u \mathbf{y}_{l-1}^t + \mathbf{H}_l^u \mathbf{s}_l^{t-1} + \mathbf{b}_l^u) \tag{84}$$
$$\mathbf{v}_l^t = \mathbf{g}(\mathbf{W}_l^v \mathbf{y}_{l-1}^t + \mathbf{H}_l^v \mathbf{s}_l^{t-1} + \mathbf{b}_l^v) \tag{85}$$
$$\mathbf{s}_l^t = \mathbf{u}_l^t \odot \mathbf{s}_l^{t-1} + (\mathbb{1} - \mathbf{u}_l^t) \odot \mathbf{h}(\mathbf{W}_l^c \mathbf{y}_{l-1}^t + \mathbf{H}_l^c (\mathbf{v}_l^t \odot \mathbf{s}_l^{t-1}) + \mathbf{b}_l^c) \tag{86}$$
$$= \mathbf{g}(\mathbf{W}_l^u \mathbf{y}_{l-1}^t + \mathbf{H}_l^u \mathbf{s}_l^{t-1} + \mathbf{b}_l^u) \odot \mathbf{s}_l^{t-1} +$$
$$(\mathbb{1} - \mathbf{g}(\mathbf{W}_l^u \mathbf{y}_{l-1}^t + \mathbf{H}_l^u \mathbf{s}_l^{t-1} + \mathbf{b}_l^u)) \odot \mathbf{h}(\mathbf{W}_l^c \mathbf{y}_{l-1}^t$$
$$+ \mathbf{H}_l^c (\mathbf{g}(\mathbf{W}_l^v \mathbf{y}_{l-1}^t + \mathbf{H}_l^v \mathbf{s}_l^{t-1} + \mathbf{b}_l^v) \odot \mathbf{s}_l^{t-1}) + \mathbf{b}_l^c), \tag{87}$$

where typically $\mathbf{g} = \boldsymbol{\sigma}$ and $\mathbf{h} = \mathbf{tanh}$. Note that we reformulated Equation 87 such that one can immediately see the resemblance to the generalized Equations 25-26 of the main text.

As one can see, a GRU does not have a separate output equation. Therefore, in order to remain consistent with the previous notation and to apply OSTL, we introduce a simple output equation

$$\mathbf{y}_l^t = \mathbf{s}_l^t, \tag{88}$$

which does not change the behavior of the GRU.

Using this formulation, the eligibility traces can be calculated according to Equation 26 as

$$\mathbf{e}_l^{t,\mathbf{W}^u} = \boldsymbol{\epsilon}_l^{t,\mathbf{W}^u} \tag{89}$$
$$\mathbf{e}_l^{t,\mathbf{H}^u} = \boldsymbol{\epsilon}_l^{t,\mathbf{H}^u} \tag{90}$$
$$\mathbf{e}_l^{t,\mathbf{b}^u} = \boldsymbol{\epsilon}_l^{t,\mathbf{b}^u} \tag{91}$$
$$\mathbf{e}_l^{t,\mathbf{W}^v} = \boldsymbol{\epsilon}_l^{t,\mathbf{W}^v} \tag{92}$$
$$\mathbf{e}_l^{t,\mathbf{H}^v} = \boldsymbol{\epsilon}_l^{t,\mathbf{H}^v} \tag{93}$$
$$\mathbf{e}_l^{t,\mathbf{b}^v} = \boldsymbol{\epsilon}_l^{t,\mathbf{b}^v} \tag{94}$$
$$\mathbf{e}_l^{t,\mathbf{W}^c} = \boldsymbol{\epsilon}_l^{t,\mathbf{W}^c} \tag{95}$$
$$\mathbf{e}_l^{t,\mathbf{H}^c} = \boldsymbol{\epsilon}_l^{t,\mathbf{H}^c} \tag{96}$$
$$\mathbf{e}_l^{t,\mathbf{b}^c} = \boldsymbol{\epsilon}_l^{t,\mathbf{b}^c}, \tag{97}$$

using Equation 10 the eligibility vectors in Equations 89-97 become

$$\epsilon_l^{t,\mathbf{W}^u} = \frac{d\mathbf{s}_l^t}{d\mathbf{s}_l^{t-1}}\epsilon_l^{t-1,\mathbf{W}^u} + \mathrm{diag}\left(\mathbf{s}_l^{t-1} \odot \mathbf{g'}_l^{u,t}\right)\mathbf{\Upsilon}_{l-1}^t - \mathrm{diag}\left(\mathbf{h}_l^{s,t} \odot \mathbf{g'}_l^{u,t}\right)\mathbf{\Upsilon}_{l-1}^t \tag{98}$$

$$\epsilon_l^{t,\mathbf{H}^u} = \frac{d\mathbf{s}_l^t}{d\mathbf{s}_l^{t-1}}\epsilon_l^{t-1,\mathbf{H}^u} + \mathrm{diag}\left(\mathbf{s}_l^{t-1} \odot \mathbf{g'}_l^{u,t}\right)\mathbf{\Upsilon}_l^{t-1} - \mathrm{diag}\left(\mathbf{h}_l^{s,t} \odot \mathbf{g'}_l^{u,t}\right)\mathbf{\Upsilon}_l^{t-1} \tag{99}$$

$$\epsilon_l^{t,\mathbf{b}^u} = \frac{d\mathbf{s}_l^t}{d\mathbf{s}_l^{t-1}}\epsilon_l^{t-1,\mathbf{b}^u} + \mathrm{diag}\left(\mathbf{s}_l^{t-1} \odot \mathbf{g'}_l^{u,t}\right) - \mathrm{diag}\left(\mathbf{h}_l^{s,t} \odot \mathbf{g'}_l^{u,t}\right) \tag{100}$$

$$\epsilon_l^{t,\mathbf{W}^v} = \frac{d\mathbf{s}_l^t}{d\mathbf{s}_l^{t-1}}\epsilon_l^{t-1,\mathbf{W}^v} + \mathrm{diag}\left((\mathbb{1} - \mathbf{u}_l^t) \odot \mathbf{h'}_l^{s,t}\right)\mathbf{H}_l^c \mathrm{diag}\left(\mathbf{s}_l^{t-1} \odot \mathbf{g'}_l^{v,t}\right)\mathbf{\Upsilon}_{l-1}^t \tag{101}$$

$$\epsilon_l^{t,\mathbf{H}^v} = \frac{d\mathbf{s}_l^t}{d\mathbf{s}_l^{t-1}}\epsilon_l^{t-1,\mathbf{H}^v} + \mathrm{diag}\left((\mathbb{1} - \mathbf{u}_l^t) \odot \mathbf{h'}_l^{s,t}\right)\mathbf{H}_l^c \mathrm{diag}\left(\mathbf{s}_l^{t-1} \odot \mathbf{g'}_l^{v,t}\right)\mathbf{\Upsilon}_l^{t-1} \tag{102}$$

$$\epsilon_l^{t,\mathbf{b}^v} = \frac{d\mathbf{s}_l^t}{d\mathbf{s}_l^{t-1}}\epsilon_l^{t-1,\mathbf{b}^v} + \mathrm{diag}\left((\mathbb{1} - \mathbf{u}_l^t) \odot \mathbf{h'}_l^{s,t}\right)\mathbf{H}_l^c \mathrm{diag}\left(\mathbf{s}_l^{t-1} \odot \mathbf{g'}_l^{v,t}\right) \tag{103}$$

$$\epsilon_l^{t,\mathbf{W}^c} = \frac{d\mathbf{s}_l^t}{d\mathbf{s}_l^{t-1}}\epsilon_l^{t-1,\mathbf{W}^c} + \mathrm{diag}\left((\mathbb{1} - \mathbf{u}_l^t) \odot \mathbf{h'}_l^{s,t}\right)\mathbf{\Upsilon}_{l-1}^t \tag{104}$$

$$\epsilon_l^{t,\mathbf{H}^c} = \frac{d\mathbf{s}_l^t}{d\mathbf{s}_l^{t-1}}\epsilon_l^{t-1,\mathbf{H}^c} + \mathrm{diag}\left((\mathbb{1} - \mathbf{u}_l^t) \odot \mathbf{h'}_l^{s,t}\right)\mathbf{P}_l^{t-1} \tag{105}$$

$$\epsilon_l^{t,\mathbf{b}^c} = \frac{d\mathbf{s}_l^t}{d\mathbf{s}_l^{t-1}}\epsilon_l^{t-1,\mathbf{b}^c} + \mathrm{diag}\left(\mathbb{1} - \mathbf{u}_l^t\right) \odot \mathbf{h'}_l^{s,t}, \tag{106}$$

with

$$\frac{d\mathbf{s}_l^t}{d\mathbf{s}_l^{t-1}} = \mathrm{diag}(\mathbf{u}_l^t) + \mathrm{diag}\left(\mathbf{s}_l^{t-1} \odot \mathbf{g'}_l^{u,t}\right)\mathbf{H}_l^u - \mathrm{diag}\left(\mathbf{h}_l^{s,t} \odot g'_l^{u,t}\right)\mathbf{H}_l^u$$
$$+ \mathrm{diag}\left((\mathbb{1} - \mathbf{u}_l^t) \odot \mathbf{h'}_l^{s,t}\right)\mathbf{H}_l^c \mathrm{diag}(\mathbf{v}_l^t)$$
$$+ \mathrm{diag}\left((\mathbb{1} - \mathbf{u}_l^t) \odot \mathbf{h'}_l^{s,t}\right)\mathrm{diag}(\mathbf{s}_l^{t-1})\mathbf{H}_l^v \mathrm{diag}(\mathbf{s}_l^{t-1} \odot \mathbf{g'}_l^{v,t})\mathbf{H}_l^c \tag{107}$$

We used the short-hand notation of $\frac{d\psi_l^t}{d\omega^t} = \mathbf{g'}|_{\psi_l^t} = \mathbf{g'}_l^{\psi,t}$ with $\psi = \{\mathbf{i}, \mathbf{v}, \mathbf{f}, \mathbf{s}\}$ and $\boldsymbol{\omega} = \{\mathbf{y}_l^{t-1}, \boldsymbol{\theta}_l\}$, $\mathbf{h}(\mathbf{W}_l^c \mathbf{y}_{l-1}^t + \mathbf{H}_l^c(\mathbf{g}(\mathbf{W}_l^v \mathbf{y}_{l-1}^t + \mathbf{H}_l^v \mathbf{s}_l^{t-1} + \mathbf{b}_l^v) \odot \mathbf{s}_l^{t-1}) + \mathbf{b}_l^c) = \mathbf{h}_l^{s,t}$ and $\frac{d\mathbf{h}_l^{s,t}}{d\boldsymbol{\omega}} = \mathbf{h'}|_{\mathbf{W}_l^c \mathbf{y}_{l-1}^t + \mathbf{H}_l^c(\mathbf{g}(\mathbf{W}_l^v \mathbf{y}_{l-1}^t + \mathbf{H}_l^v \mathbf{s}_l^{t-1} + \mathbf{b}_l^v) \odot \mathbf{s}_l^{t-1}) + \mathbf{b}_l^c} = \mathbf{h'}_l^{s,t}$. Furthermore, the elements of $\mathbf{\Upsilon}_l^t$ are given by $(\mathbf{\Upsilon}_l^t)_{opq} = \delta_{op}(y_l^t)_q$ and the elements of $\mathbf{P}_l^t$ are given by $(P_l^t)_{opq} = \delta_{op}(r_l^{t+1} s_l^t)_q$

Considering the mean squared error loss with readout weights given in Equation 17 of the main text

$$E^t = \frac{1}{2}\sum_{n_k}\left(\mathbf{W}^{\mathrm{out}}\mathbf{y}_k^t - \hat{\mathbf{y}}^t\right)^2, \tag{108}$$

the learning signals become

$$\mathbf{L}_l^t = \mathbf{L}_{l+1}^t \left(\mathrm{diag}\left(\mathbf{s}_l^{t-1} \odot \mathbf{g'}_l^{u,t}\right)\mathbf{W}_{l+1}^u\right.$$
$$\left.-\mathrm{diag}\left(\mathbf{h}_l^{s,t} \odot \mathbf{g'}_l^{u,t}\right)\mathbf{W}_{l+1}^u + \mathrm{diag}\left((\mathbb{1} - \mathbf{u}_l^t) \odot \mathbf{h'}_l^{s,t}\right)\left[\mathbf{W}_{l+1}^c + \mathbf{H}_{l+1}^c \mathrm{diag}\left(\mathbf{g'}_l^{v,t}\right)\mathbf{W}_{l+1}^v\right]\right), \tag{109}$$

with

$$\mathbf{L}_k^t = \left(\left(\mathbf{W}^{\mathrm{out}}\right)^T \mathbf{y}_k^t - \hat{\mathbf{y}}^t\right). \tag{110}$$

## Supplementary material S7: Simulation setup

The simulations were performed on a standard desktop PC with a single NVIDIA RTX2080 GPU, while the hyperparameter exploration was performed on a supercomputer cluster. For the neural network implementation we used the TensorFlow deep learning framework, version 1.10.0, with GPU support enabled.

**Music prediction**

We used the JSB dataset [4] with the standard training/testing data split, i.e., 229 chorales for training and 77 chorales for testing. We employed an architecture comprising a single RNN layer with 150 units and a stateless output layer comprising 88 sigmoid units, as in [5]. In particular, we analyzed three types of RNN layers, a feed-forward SNU layer configured to $\mathbf{h} =$Step, $\mathbf{g} = \mathbb{1}$ and $d = 0.4$, a feed-forward sSNU layer configured to $\mathbf{h} = \boldsymbol{\sigma}$, $\mathbf{g} = \max(0, x) = \mathbf{ReL}$ and $d = 0.8$, as well as an LSTM layer. We used the standard stochastic gradient descent optimizer and trained with BPTT, OSTL, OSTL rand and random e-prop for at least 100 epochs, or until the test performance did not improve for more than 15 epochs. In OSTL rand, we used a fixed random feedback alignment to deliver the learning signals to the RNN layer, see Supplementary material S3. The fixed random feedback matrix in OSTL rand as well as in random e-prop was initialized with random numbers according to a univariate Gaussian distribution with mean $\mu = 0$ and variance $\mathrm{var} = 1$. The assessment was done based on the common approach for JSB to report the negative log-likelihood. For OSTL we investigated learning rates between $10^{-5} - 10^{-2}$. Our final results were obtained using learning rates of $0.0005$, $0.0001$ and $0.0003$ for the SNU, sSNU and LSTM network, respectively. To ensure repeatability of the results, we averaged them over 5 networks trained with different random initial conditions.

**Digit classification**

We used the MNIST dataset [6] with the standard training/testing data split, i.e., 60000 examples for training and 10000 examples for testing. The grey values of the pixels were transformed into spike rates using the same procedure as presented in [5] with $N_s = 20$. We employed a deep feed-forward architecture of two layers of sSNUs with 256 units each and an output layer of 10 sSNUs, all configured to $\mathbf{h} = \boldsymbol{\sigma}$, $\mathbf{g} = \mathbf{LReL}$ (leaky rectified linear function) with leak $\alpha = 0.01$, and $d = 0.9$. In addition, we employed a feed-forward architecture of two layers of SNUs with 256 units each and an output layer of 10 SNUs, all configured to $\mathbf{h} = \mathbf{Step}$, $\mathbf{g} = \mathbf{LReL}$ with $\alpha = 0.01$ and $d = 0.9$. We used the standard stochastic gradient descent optimizer and trained for 100 epochs with BPTT and OSTL. The assessment was done based on the classification accuracy. For OSTL we investigated learning rates between $10^{-3} - 0.5$. Our final results were obtained using learning rates of $0.15$ and $0.2$ for the sSNU and the SNU configuration, respectively. To ensure repeatability of the results, we averaged them over 5 networks trained with different random initial conditions.

**Language modelling**

The PTB dataset [7] in a word-level version preprocessed by Tomáš Mikolov [8] was used for language modelling. It comprises sequences of words corresponding to English sentences from the Wall Street Journal, restricted to the most frequent 10k tokens and split into training, validation and test subsequences with around 930k, 74k and 82k tokens, respectively. We employed an architecture comprising an embedding layer with 10000 units, followed by three feed-forward sSNU layers with 1300 units each, configured to $\mathbf{h} = \boldsymbol{\sigma}$, $\mathbf{g} = \mathbb{1}$ and $d = 0.4$, and a softmax layer with 10000 units as the output layer. Note that the weights of the softmax were tied to the input embeddings. We used the standard stochastic gradient descent optimizer with initial learning rates of $1.0$ for BPTT and in the range of $10^{-2} - 1.0$, respectively. For BPTT we used a truncation horizon of $T = 20$ words.

Our final results for OSTL were obtained with an initial learning rate of $0.08$. The initial learning rates were multiplied by a factor of $0.5$ at each epoch if the validation perplexity did not improve by more than 10%. We used gradient clipping with a ratio of $3.5$. These methods for preventing overfitting seemed effective. This can be observed in Supplementary Table 2, which shows the training perplexities along with the test perplexities. We trained for at least 20 epochs, or until the validation perplexity did not further decrease, and averaged over 5 random initializations.

Table 2: Extended performance comparison for the PTB dataset.

| UNIT | ALGORITHM | TRAIN PPL | TEST PPL |
|---|---|---|---|
| sSNU | BPTT | $87.3\mathrm{ppl} \pm 4.4$ | $132.3\mathrm{ppl} \pm 1.3$ |
| sSNU | OSTL | $90.8\mathrm{ppl} \pm 1.1$ | $137.9\mathrm{ppl} \pm 0.3$ |

**Speech recognition**

We considered an application to speech recognition based on the TIMIT dataset [9]. To this end, we used the same data preprocessing as stated in [10], but instead of 61 output classes we used the reduced set of 39 output classes. We employed an architecture comprising two layers of LSTMs with 200 units each and a stateless softmax layer with 39 units on top. In addition, we also employed an architecture comprising two layers of sSNUs configured to $\mathbf{h} = \boldsymbol{\sigma}$, $\mathbf{g} = \mathbb{1}$ and $d = 0.3$ with 200 units each and a stateless softmax layer with 39 units on top. We used the RMSProp optimizer with a decay of $0.95$ and an epsilon of $1^{-8}$ and trained for $24$ epochs with BPTT, TBPTT and OSTL. For TBPTT we analyzed the performance for truncation horizons $T \in \{10, 30, 50, 90\}$.

The assessment was done based on the phoneme error rate. For OSTL we investigated learning rates between $10^{-5} - 10^{-2}$. Our final results were obtained using learning rates of $0.00076$ for LSTMs and $0.00083$ for sSNUs, respectively.

In addition, we trained a network with the same hyperparameters as outlined above, but with just one hidden layer with OSTL, OSTL w/o $\mathbf{H}$ and with random e-prop. In OSTL w/o $\mathbf{H}$, the influence of the recurrent matrix $\mathbf{H}$ for the eligibility traces is neglected, see Supplementary material S3. For random e-prop, the fixed random feedback matrix was initialized with random numbers according to a univariate Gaussian distribution with mean $\mu = 0$ and variance $\text{var} = 1$.

To ensure repeatability, we averaged all results over 5 networks trained with different random initial conditions.



\end{document}